\documentclass[conference]{IEEEtran}
\IEEEoverridecommandlockouts
\usepackage{cite}
\usepackage{amsmath,amssymb,amsfonts}
\usepackage{graphicx}
\usepackage{textcomp}
\usepackage{xcolor}
\usepackage{hyperref}
\usepackage{url}
\usepackage{bbm}
\usepackage{balance}
\usepackage{booktabs}       
\usepackage{amsfonts}       
\usepackage{nicefrac}       
\usepackage{microtype}      
\usepackage{xcolor}

\usepackage{array}
\usepackage{tabu}
\usepackage{multirow}
\usepackage{graphicx}
\usepackage{amsmath}
\usepackage[ruled,linesnumbered]{mathtools,algorithm2e}
\usepackage[noend]{algpseudocode}
\usepackage{wrapfig} 
\RequirePackage{expl3}

\usepackage{multicol}

\usepackage{enumitem}
\usepackage{xcolor}
\usepackage{caption}
\usepackage{graphicx}
\usepackage{float} 
\usepackage{subfigure}
\usepackage{subcaption}
\usepackage{pgfplots}

\usepackage{longtable}

\usepackage{pifont}
\usepackage{threeparttable}

\makeatletter
\newcommand{\myonecolumn}{%
    \global\columnwidth\textwidth
    \global\hsize\columnwidth
    \global\linewidth\columnwidth
    \global\@twocolumnfalse
    \global\@columnwidth\columnwidth
    \global\@hsize\columnwidth
    \global\@linewidth\columnwidth
    \global\@totalleftmargin 0pt%
    \global\@rightmargin 0pt%
    \global\@restonecolfalse
    \@normalsize
}
\makeatother

\def\BibTeX{{\rm B\kern-.05em{\sc i\kern-.025em b}\kern-.08em
    T\kern-.1667em\lower.7ex\hbox{E}\kern-.125emX}}
    
\begin{document}

\title{Rank Supervised Contrastive Learning for Time Series Classification\\
}

\author{\IEEEauthorblockN{Qianying Ren}
\IEEEauthorblockA{\textit{School of Computing} \\
\textit{University of Connecticut}\\
Storrs, United States \\
qianying.ren@uconn.edu}
\and
\IEEEauthorblockN{Dongsheng Luo}
\IEEEauthorblockA{\textit{Knight Foundation School of Computing and Information Sciences} \\
\textit{Florida International University}\\
Florida, United States \\
dluo@fiu.edu}
\and
\IEEEauthorblockN{\textsuperscript{*} Dongjin Song}
\IEEEauthorblockA{\textit{School of Computing} \\
\textit{University of Connecticut}\\
Storrs, United Statse\\
dongjin.song@uconn.edu}
}

\maketitle
\begin{abstract}
Recently, various contrastive learning techniques have been developed to categorize time series data and have exhibited promising performance for real-world applications. A general paradigm is to utilize appropriate data augmentation methods and construct feasible positive samples such that the encoder can yield robust and discriminative representations by mapping similar data points closer together in the feature space while pushing dissimilar data points farther apart. Despite its efficacy, the fine-grained relative similarity (\textit{e.g.}, rank) information of positive samples is not fully exploited, especially when labeled samples are limited. To this end,  we present Rank Supervised Contrastive Learning (RankSCL) to perform time series classification. Different from conventional contrastive learning frameworks, RankSCL augments raw data in a targeted manner in the embedding space and selects more informative positive and negative pairs for the targeted sample. Moreover, a novel rank loss is developed to assign higher weights to more confident positive pairs and lower weights to less confident positive pairs, enabling the encoder to extract the same class's fine-grained information and produce a clear boundary among different classes. Thoroughly empirical studies on 128 UCR datasets and 30 UEA datasets demonstrate that the proposed RankSCL can achieve state-of-the-art performance compared to existing baseline methods.
\end{abstract}


\section{Introduction}
Nowadays, time series data are becoming ubiquitous in numerous real-world applications. For instance, in a power plant \cite{Prickett2011}, a large number of sensors can be employed to monitor the operation status in real time. With a fitness tracking device, a temporal sequence of actions \cite{Parkka2006}, \textit{e.g.}, walking for 5 minutes, running for 1 hour, and sitting for 15 minutes, \textit{etc}, can be recorded and detected with related sensors. In healthcare, we can detect epileptic seizures by classifying EEG data into two categories, \textit{i.e.}, normal and abnormal, and provide timely medical intervention. With the huge amount of time series data, how to categorize and interpret the status becomes a critical issue to investigate.

Traditionally, one of the most popular time series classification approaches is to use the nearest neighbor (NN) classifier based on a distance measure~\cite{LinesBagnall2015}. For instance, Dynamic Time Warping (DTW) distance has been used together with an NN classifier (DTW-NN) to provide a strong baseline \cite{Bagnall2017TimeSeriesBakeOff}. In addition, ensemble methods have been shown to outperform DTW-NN by ensembling individual NN classifiers with different distance measures over the same or different feature spaces. More recently, Collective Of Transformation-based Ensembles (COTE) combines the strengths of multiple approaches to handle various aspects of time series data can yield better classification accuracy. Lines et al. further extended COTE to create HIVE-COTE \cite{Lines2018HIVECOTE} by incorporating a hierarchical vote system. These approaches, however, involve high complexity for both training and inference.

More recently, deep learning based time series classification algorithms, \textit{e.g.}, InceptionTime~\cite{InceptionTime}, are becoming more popular as they have shown promising performance and can obtain more effective representations. However, supervised learning may require a substantial amount of high-quality labeled time series data for training which could be infeasible in many real-world applications. Therefore, several contrastive learning based time series representation techniques have been developed to resolve this issue. The key idea is to leverage appropriate augmentations and construct feasible positive samples such that the encoder can yield robust and discriminative representations by mapping similar data points closer together in the feature space while pushing dissimilar data points farther apart. For instance, TimCLR \cite{timeCLR} uses the DTW \cite{DTW} to provide phase-shift and amplitude-change augmentations to make the representation learning tied to temporal variations. Despite their efficacy, the fine-grained relative similarity (\textit{e.g.}, rank) information of positive samples is not fully exploited, especially when labeled samples are limited.

To this end, we propose Rank Supervised Contrastive Learning (RankSCL) to tackle this issue and yield more effective representations to facilitate time series classification. The key idea is to rank the importance of different positive samples so as to better understand the potential landscape of feature space. Specifically, we make full use of the information of positive samples by leveraging their relatively similarity information in terms of rank. We encode the rank by taking account of the number of triplets in which the distance of anchor-negative pairs is smaller than anchor-positive pairs. Based on that, a targeted data augmentation technique is designed to generate designated samples, aiming to enrich the information of samples from the same category and enhance the boundary among samples from different categories. By combining these two techniques, our proposed RankSCL has been thoroughly evaluated on 128 UCR datasets and 30 UEA datasets. Our experiment results demonstrate that the proposed RankSCL can achieve state-of-the-art performance compared to existing baseline methods. Our main contributions include:
\begin{itemize}[leftmargin=*]
    \item We develop a novel rank supervised contrastive learning framework and present a novel rank loss that assigns different weights to different levels of positive samples.
    \item We propose a targeted data augmentation technique based on RSCL to generate designated positive samples that can enrich the information of samples from the same category and enhance the boundary among different categories.
    \item Our empirical studies on 128 UCR datasets and 30 UEA datasets demonstrate that the proposed RankSCL outperforms the state-of-the-art.
\end{itemize}
\section{Related Work}
The proposed RankSCL is closely related to contrastive time series representation learning, time series classification, and time series based data augmentation.

\subsection{Contrastive Time Series Representation Learning}
For contrastive learning, it is essential to enrich the representation space through the manual generation of positive and negative pairs. In traditional, positive pairs representations are aligned closely together, while negative pairs are deliberately distanced.  As illustrated by the SimCLR \cite{simclr} framework, different views of augmentations for the same sample are treated as positive pairs, while augmentations applied to different samples are regarded as negative pairs. However, recent works in time series contrastive learning have introduced various pair designs to leverage the invariant features of time series data. For instance, TimCLR \cite{timeCLR} employs the DTW \cite{DTW} to facilitate phase-shift and amplitude-change augmentations, which are better suited for time series context, to ensure representation learning is intrinsically connected to temporal variations. The contrastive loss is further defined in both instance-wise and patch-wise levels by TS2Vec \cite{ts2vec} framework, which separates several time series into various patches. Meanwhile, a novel temporal contrastive learning task is introduced by TS-TCC \cite{tstcc}, prompting augmentations to predict each other's future sequence. To capture distinctive seasonal and trend patterns, CoST \cite{Cost} applies contrastive losses across time and frequency domains. Additionally, TF-C \cite{TF-C} aims to optimize time-based and frequency-based representations of the same example to be closer and introduces an innovative time-frequency consistency framework. InfoTS \cite{InfoTS} leverages information-theoretic principles to generate appropriate time series data augmentations by maximizing high fidelity and variety. Despite these methodological innovations, they seldom leverage the fine-grained relative similarity (e.g., rank) information of
positive samples.


\subsection{Time Series Classification}
The field of time series classification is fundamental and rapidly evolving. Some conventional methods are not based on deep learning such as TS-CHIEF \cite{shifaz2020ts}, ROCKET \cite{ROCKET}, and DTW-NN \cite{DTW-NN}, have been foundational. With the rapid development of deep learning, an increasing number of models are emerging in this domain, exhibiting remarkable performance. Deep learning algorithms can get better classification results than statistics-based methods because they are better at learning rich representations during training. InceptionTime\cite{InceptionTime} applied the Inception Networks in time series and can capture local patterns. The attention mechanism was employed by MACNN \cite{mcnn} to enhance the multi-scale CNNs' classification performance. To address multi-variate TSC difficulties, Hao et al. introduced CA-SFCN \cite{ca-sfcn}, which used variable and temporal attention modulation. 

\subsection{Time Series Data Augmentation}
Data augmentation is essential for the successful use of deep learning models on time series data since it is a powerful technique to increase the quantity and size of the training data. Conventional methods of time series data enhancement can be broadly categorized into three domains, the first method is the time domain, the second is the frequency domain, and the third is the time-frequency domain. The easiest data augmentation techniques for time series data are the transforms in the time domain. The majority of them perform direct manipulations on the original input time series, such as adding Gaussian noise or more intricate noise patterns like spikes, steps, and slopes. However, there are a few ways to consider data augmentation from a frequency domain view. For time-frequency domain, Yao et al. \cite{STFNets} use the short Fourier transform(STFT) to create time-frequency features for sensor time series. They then supplement the data by using these features so that a deep LSTM neural network can classify human activities. 

\begin{figure*}[!h] 
\centering 
\includegraphics[width=0.99\textwidth]{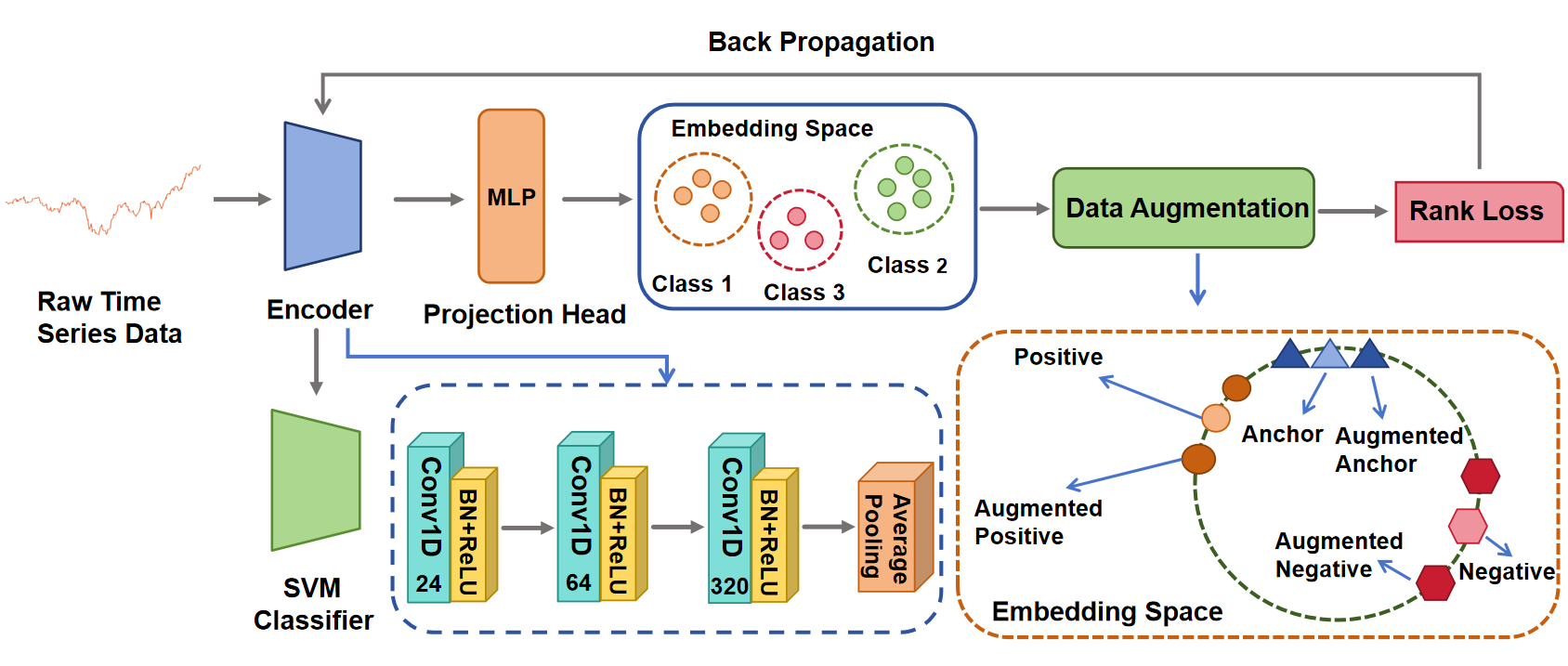} 
\caption{Overview of RankSCL framework, consisting of three components: (1) a Fully Convolutional Network that captures the embeddings of raw time series instances, (2) targeted data augmentation that generates more samples in the embedding space, (3) selection of valid triplets and calculation of rank loss to train the encoder network. Even though this figure shows a univariate time series instance as an example, the architecture supports multivariate instances.} 
\label{Fig_main} 
\end{figure*}

\section{Method}

\subsection{Notations and Problem Definition}
A time series instance $x_{i}$ is represented by a $T\times F$ matrix, where $T$ is the time step and $F$ is the feature dimension. With $F=1$, $x$ is a univariate instance, otherwise $x$ is known as a multivariate instance. Given a set of $N$ time series instances $\mathbbm{X}=\{x_1,x_2,x_3,...,x_N\}$, the objective is to learn a nonlinear function $f_\theta$ that maps each $x$ to a $D$ dimensional vector $\mathbf{v}\in \mathbbm{R}^D$, which preserves its semantics and $D  \ll T\times F$. In supervised settings, we have a subset of $\mathbbm{X}$, denoted by $\mathbbm{X}_L$, where each instance $x$ is associated with a label $y$. 

\subsection{Framework}


The overall architecture is shown in Figure~\ref{Fig_main}. We feed the raw time series data into an \textit{encoder network}, $f_{\theta}$, to learn low-dimensional representations, which are then forward propagation into a \textit{projection head}. For each sample, we treat it as an anchor and define the positive samples based on the label information \cite{khosla2020supervised}. Specifically, instances with the same label as the anchor are considered positive samples. Otherwise, they are used as negative samples. To enrich the intra-class information, we utilize data augmentation techniques on embeddings of all samples. A rank-supervised contrastive loss function is adopted to train the encoder network and projection head. To infer representations for time series classification, we follow the existing contrastive learning frameworks \cite{simclr} to ignore the projection head and use hidden representations produced by the encoder network.

\subsection{Model Architecture.} 

The main components of our method contain the encoder network and the projection head.

\noindent \textbf{Encoder Network, $f_{\theta}(\cdot)$}. As shown in Figure~\ref{Fig_main}, we utilize 3-layer Fully Convolutional Networks (FCN) to map the input raw time series data $x_{i}$ in the representation $r_{i} = f_{\theta}(x_{i})$.
Each module consists of a convolutional layer followed by a batch normalization layer \cite{batch_norm} and a Rectified Linear Units (ReLU) activation function \cite{ReLU}. 
To minimize the number of weights, features are routed into a global average pooling layer after the convolution blocks.

\noindent \textbf{Projection Head, $g_{\theta}(\cdot)$}. 
Inspired by the classic SimCLR framework \cite{simclr}, we included a small MLP network as a projection head to further transform $r$ to a new space that $z_{i} = g_{\theta}(r_{i})$.  We adopt a normalization function following the last layer to map representations in a unit hypersphere to facilitate the distance and ranking computation.  The projection head is only used in the training phase and will be dropped in the inference. 

\subsection{Data Augmentation on the Embedding Space}
In the paradigm of contrastive learning,  data augmentation emerges as a pivotal component, significantly influencing the effectiveness of the model. The selection of a suitable augmentation strategy, particularly for time series data, stands as a crucial challenge. In the literature, a common approach involves selecting augmented data points that are substantially divergent from the original data, thereby introducing increased variability to foster a more robust encoder. Yet, this strategy often leads to issues of distributional shifts, as highlighted in recent studies \cite{ts2vec}. 

Diverging from traditional contrastive learning approaches that apply data augmentation directly to raw time series data, our method addresses these challenges within the embedding space. For each training instance and its corresponding label  $(x_i,y_i)$ in the labeled dataset  $\mathbbm{X}_L$, we utilize the encoder network coupled with a projection head to generate a compact embedding $z_i$ of $x_i$ within a unit hypersphere. We introduce jittering with scale $\alpha$, denoted by $t_\alpha$  as a selected augmentation operation. An augmented instance is then obtained through $z_i' = t_\alpha(z)$.  

To further enhance diversity, we consider a set of two jittering operations with distinct scales, $\mathcal{T}=\{t_{\alpha_1},t_{\alpha_2}\}$. In practice, we set $\alpha_1 = 0.03$ and $\alpha_2 = 0.05$. As indicated in prior research \cite{ts2vec}, augmentations performed in the hidden space effectively preserve label information. Consequently, both the augmented instances and their original labels are incorporated into the current training batch.

By applying small-scale jittering to embeddings in a compact unit hypersphere, our augmentation technique not only enriches valid information but also mitigates potential issues related to distribution drift or the creation of outliers. This approach, in turn, enhances the training process of the encoder network within the contrastive learning framework. 

\subsection{Selection of Valid Triplet Pairs}
In the realm of contrastive learning, the foundational goal is to maximize the similarity between positive samples and ensure that negative samples remain distinctly separate. Early contrastive learning models typically incorporated a single positive and a negative pair within a minibatch, as established by seminal works \cite{chopra2005learning}. Recent advancements have introduced multiple positive and negative pairs \cite{sohn2016improved,khosla2020supervised}, leading to significant strides in diverse fields including computer vision \cite{khosla2020supervised} and natural language processing \cite{gunel2021supervised}. However, these methods often involve considering all anchor-negative pairs, which can be computationally intensive.

To optimize this process, we propose the novel concept of a ``valid hard negative pair'' for contrastive learning. In this context, for an anchor instance $x_a$ and a corresponding positive sample 
$x_p$, a negative sample $x_n$ is deemed ``valid hard'' if the distance $dist(x_a,x_n)$ between the anchor $x_a$ and the negative sample $x_n$ is less than the distance between the anchor and the positive sample, i.e., $dist(x_a,x_n) < dist(x_a,x_p)$. A triplet formed by these criteria, $(x_a,x_p,x_n)$, is then defined as a \textit{valid triplet pair}. This approach focuses on more challenging and informative negative examples during training, thereby enhancing the discriminative power of the resultant representations.

The introduction of valid triplet pairs significantly enhances class separation while maintaining proximity between positive samples and their anchors. Additionally, focusing selectively on valid triplets during contrastive training emphasizes more informative negative samples, reducing computational demands and thereby improving the algorithm's overall efficiency.

\begin{figure}[t] 
\centering 
\includegraphics[width=0.4\textwidth]{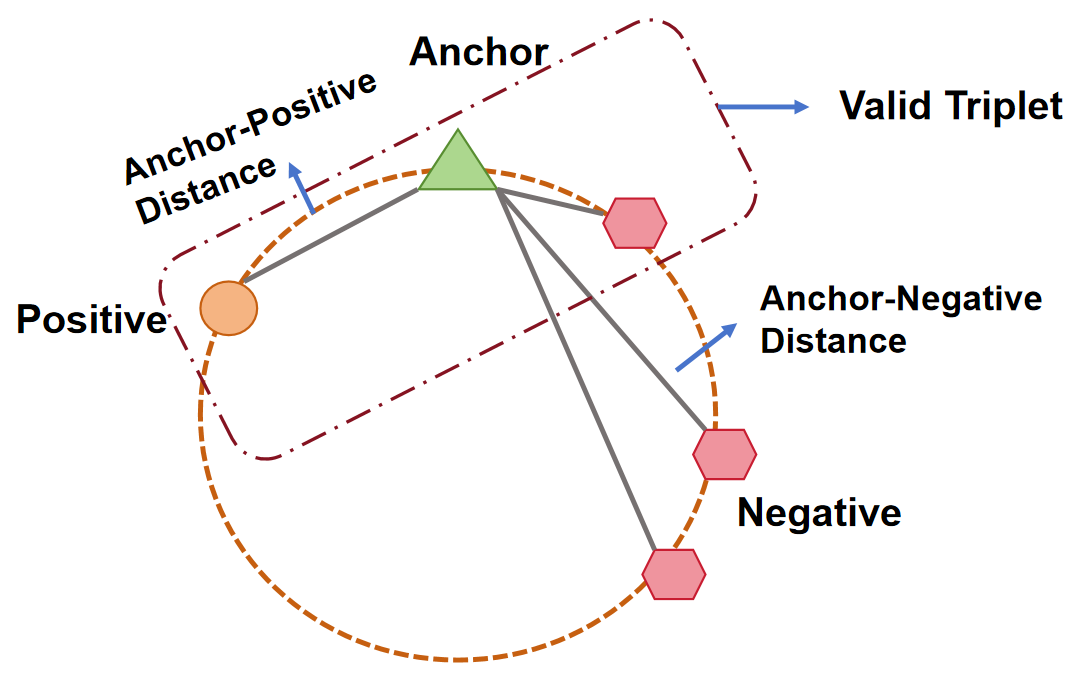} 
\caption{Valid Triplet Selection} 
\label{mining} 
\end{figure}


\subsection{Rank Supervised Contrastive Learning Loss}
In traditional contrastive learning, a key overlooked aspect is the variable distances of different positive samples from the anchor. Treating all positive samples equally, without considering their proximity to the anchor, may not effectively capture the nuanced potential representations of a class. Additionally, this approach is susceptible to the influence of outliers, potentially introducing noise into the learning process. 

To address this issue, we introduce the concept of ranking different positive samples and propose a novel rank-supervised contrastive learning loss function. This function differentially weights positive samples based on their utility in training the model. We first define the set of valid hard negative samples for an anchor-positive pair $(x_a,x_p)$ as:
\begin{equation}
    \mathbbm{X}_{ap}^{(n)} = \{ x_n | dist(x_a,x_n) < dist(x_a,x_p)\}
\end{equation}

The intuition behind this approach is that a positive sample with fewer valid hard negative samples is likely closer to the anchor and therefore more informative for the class representation. Such samples should be given more weight. Conversely, a positive sample with a larger $\mathbbm{X}_{ap}^{(n)}$  might be distant from the class centroid and treated with lower weight, as it could potentially be an outlier introducing noise. To implement this, we rank positive samples based on the count of their valid hard negative samples. In this framework, distinct positive samples receive different weights during the learning process. A high-rank positive sample, indicating a larger number of valid hard negatives, may offer less valuable information for the class and is more likely to be an outlier. In contrast, a low-rank positive sample, with fewer valid hard negatives, is weighted more heavily, suggesting it provides more relevant information for the class.

Formally, we define the rank of a pair $(x_a,x_p)$ based on the size of its valid hard negative sample set:
\begin{equation}
\label{eq:1}
R(x_a,x_p)=\sum_{n}^{|\mathbbm{X}_{ap}^{(n)}|}\mathbbm{1}(dist(x_a,x_n)\leq dist(x_a,x_p)).
\end{equation}
Here, $\mathbbm{1}(\cdot)$ is an indicator function, returning 1 if the condition is met, otherwise 0. This rank-based approach aims to enhance the model's ability to discern between more and less informative positive samples, thereby refining the training process and improving the overall quality of learned representations.
\begin{figure}[t] 
\centering 
\includegraphics[width=0.4\textwidth]{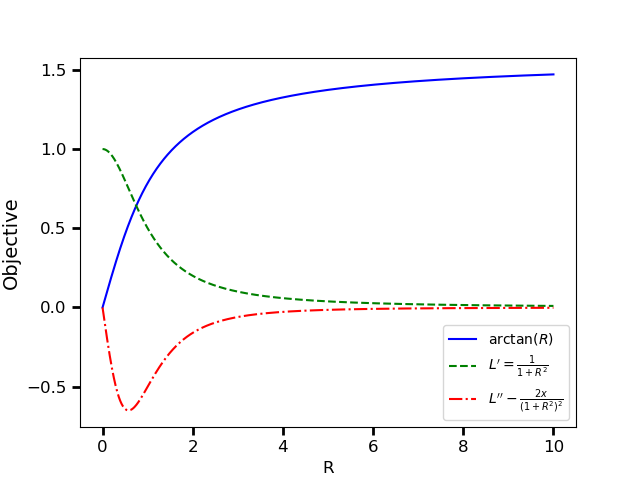} 
\caption{Properties of rank function $R(\cdot)$} 
\label{rankingloss} 
\end{figure}

The discrete nature of the indicator function in the rank function $R(x_a,x_p)$  presents a challenge for optimization during the training process. To address this, we replace the indicator function with a continuous approximation using the sigmoid function $\sigma(\cdot)$. Consequently, the revised rank function becomes:
 \begin{equation}
R(x_a,x_p)=\sum_{n}^{|\mathbbm{X}_{ap}^{(n)}|}\sigma(dist(x_a,x_p)-dist(x_a,x_n)) \label{eq:2},
\end{equation}
where $\sigma(k)=\frac{1}{1+e^{-k}}$ is the sigmoid function.

In order to assign differentiated weights to positive samples based on their ranks, we introduce a novel objective function that penalizes poorly ranked samples more stringently. Samples with lower ranks, indicating a greater number of closer negative samples to the anchor, should receive lesser weight. This is because these samples may potentially be outliers, providing noisy information that is not conducive for extracting general class representations. For this purpose, we employ the arctan function to map the ranks. This function ensures that as $R$ increases linearly, the increment in the loss function gradually diminishes, reflecting the decreasing significance of higher-ranked (potentially noisier) samples. Therefore, our final objective function is:
 \begin{equation}
\mathcal{L}(R(x_a,x_p)) = \sum_{a}\sum_{p}\arctan(R(x_a,x_p)) \label{eq:3}
\end{equation}
This loss function can then be used to perform backpropagation during the training process. Utilizing the Adam optimizer, this methodology facilitates the learning of discriminative representations, effectively balancing the need to prioritize informative samples while mitigating the impact of less useful outliers.

\noindent \textbf{Remark.} The choice of mapping functions significantly influences the results of our model. In exploring different functions, we compared the performance of the $\log(1+z)$ function against the $\arctan (x)$ function. Our findings indicated that the $\log(1+z)$ function led to inferior performance compared to the $\arctan (x)$ function. This discrepancy arises because, with an increasing rank value, the rate of change in the $\log(1+z)$ function is more pronounced than in the arctan function. Additionally,   arctan has upper and lower bounds, meaning that for higher rank values, the samples with minimal derivative changes will have a slower rate of change, thereby exerting less influence on the learning process.
The conceptual framework of our loss function design shares some similarities with the approach in \cite{song2018deep}, yet it remains distinct in application and methodology. Our method utilizes the inverse tangent function to map ranking values, applying this in the context of representation learning. This approach effectively adjusts the weights of different positive samples, enhancing the model's ability to distinguish class boundaries and extract meaningful learning representations for various categories. This nuanced application of the arctan function caters specifically to the challenges of rank-supervised contrastive learning, balancing the need for effective class separation with the nuances of representation learning. 

\subsection{Algorithms}
\label{app:theorem}


The training algorithm of RankSCL is described in Algorithm \ref{alg:al_1}. First, the parameters in the encoder $f_\theta$ and projection head $g_\theta$ are randomly initialized (line 1). For a batch of training samples $\mathbb{X_{B}} \subseteq \mathbb{X}$, we can get the representations $r_{i} = f_\theta(x_{i})$ and then get the embeddings  $z_{i} = g_\theta(r_{i})$ (lines 4-5).  Two jittering operations with distinct scales $\alpha_{1}=0.03,\alpha_2=0.05$ are applied to the normalized embeddings. All augmented samples are assigned the same labels as the original data and then we combine them to get larger samples (lines 6-11). After doing the normalization, all of the samples will be embedded in the unit hypersphere (line 12). We leverage label information to help us categorize different classes (line 13). The Euclidean distances are calculated for each pair of samples in the batch and the rank value of each anchor-positive pair is also calculated (lines 14-15). Parameters $\theta$ in the encoder and projection head are updated by minimizing the rank-supervised contrastive learning loss (line 17-18).  We will release the code upon the acceptance of this work.

\setlength\intextsep{3pt}
\begin{algorithm}[htb]

 \caption{RankSCL's main learning algorithm}
 \label{alg:al_1}
  \KwIn{time series dataset $\mathbb{X}$, the label set $\mathbb{Y}$, number of augmentation $k$ }
  \KwOut{encoder network $f_{\theta}$, projection head $g_{\theta}$}

   Initialize the encoder $f_{\theta}$, projection head $g_{\theta}$.; \\
    \For{each epoch}
    {
       
        \For{each training epoch $\mathbb{X_{B}}\subseteq 
  \mathbb{X}$}
        {
           $r_{i} \leftarrow f_\theta(x_{i})$; \\
           $z_{i} \leftarrow g_\theta(r_{i})$;\\
           \While{$j<k$}{
           $z_{i}'=t_{\alpha_1}(z_{i})$;\\
           $z_{i}''=t_{\alpha_2}(z_{i})$;\\
           $j \leftarrow j+1$;\\
           }
            $Z_{B}\leftarrow average\left( {Z_{B},Z'_{B},Z''_{B}}\right)$;\\
            Normalize $Z_{B}$;\\
            Categorize $Z_{B}$ by label set $\mathbb{Y}$;\\
            Compute Euclidean distance matrix;\\
            Compute rank $\textit{R}$ for each anchor-positive pair with Eq.(3);\\
            Compute rank-supervised contrastive learning loss $\mathcal{L}$ with Eq.(4);\\
            Update parameters $\theta$ in the encoder $f_\theta$;\\
            Update parameters $\theta$ in the projection head $g_\theta$;\\
        }
    }
\end{algorithm}

\section{Experiments}\label{sec:experiment}
We assess the learned representations of our method on time series classification task in this part. The comprehensive experimental results and details are shown in the appendix.

\begin{table*}[!htb]
\small
  \centering

    \begin{tabular}{ccccccccc}
    \toprule
      
   Dataset & RankSCL & InfoTS & TS2Vec & T-Loss & TS-TCC & DTW & TimesNet & TNC\\
  \hline
    UCR   &  \textbf{2.430}     & 3.824    & 2.465 & 3.531 & 5.508&  5.277   &  5.211 & 7.754  \\
    UEA  &  \textbf{2.433}     &   3.917 & 3.650  & 4.800 & 5.550 &  4.483  &   4.233 & 6.933  \\
    \hline
    \end{tabular}%
  \label{tab:table2}%
  \caption{Average Rank values for 128 UCR datasets and 30 UEA datasets}
\end{table*}%

\begin{table*}[!htb]

  \centering
 
  \small
    \begin{tabular}{ccccccccc}
    
    \toprule
          & \multicolumn{4}{c}{128 UCR} & \multicolumn{4}{c}{30 UEA} \\
          \cmidrule(r){2-5} \cmidrule(r){6-9}
   Method & ACC $\uparrow$ & Prec. $\uparrow$ & F1 $\uparrow$ & Recall $\uparrow$ & ACC $\uparrow$ & Prec. $\uparrow$ & F1 $\uparrow$ & Recall $\uparrow$\\
  \hline
    RankSCL  &  0.821     &   \textbf{0.817}   & \textbf{0.803} & \textbf{0.789} & \textbf{0.715}    &    \textbf{0.719}    &  \textbf{0.705}  & \textbf{0.692}     \\
    InfoTS &  0.733     &   0.723   & 0.705   & 0.688  &  0.669     &  0.672     & 0.657 & 0.643\\
    TS2Vec &    \textbf{0.822}   &   0.816    &   0.799 & 0.783 &   0.695   &   0.687    & 0.672 & 0.657 \\
    TS-TCC &    0.685   &  0.603     &  0.566   & 0.533 &    0.617   & 0.573   & 0.533 & 0.499 \\

    T-Loss &   0.782    &  0.750    &  0.743 &  0.737  &  0.581     &  0.572    &0.545 & 0.520\\
    DTW   &   0.679  &   0.672    &  0.668  & 0.646 &  0.654   &  0.645    & 0.624  & 0.605 \\
    TimesNet & 0.688 & 0.675 & 0.649 & 0.625 & 0.676 & 0.664 & 0.639 & 0.616 \\
    TNC & 0.406 & 0.305 & 0.291 & 0.279 & 0.345 & 0.311 & 0.286 & 0.265\\

    \hline
    \end{tabular}
  
   \caption{Classification Results on 128 UCR datasets and 30 UEA datasets}
   \label{tab:table1}%
\end{table*}%

\subsection{Datasets and Baselines}
The classic two different kinds of benchmark datasets are used for the evaluation\footnote{\url{https://www.cs.ucr.edu/~eamonn/time\_series\_data/}}. For the multivariate time series classificatin task we use the UEA archive \cite{UEA} consists of 30 multivariate datasets, while the UCR archive \cite{UCR} has 128 univariate time series datasets for the univariate time series classficiation task. The extensive experiments are conducted compared with other SOTAs, such as InfoTS \cite{InfoTS}, TimesNet\cite{wu2022timesnet}, TS2Vec \cite{ts2vec}, T-Loss \cite{nips2019}, TNC \cite{TNC}, TS-TCC \cite{tstcc}, TST \cite{TST} and DTW \cite{DTW}. The appendix contains the complete experiment results of our method on all datasets. To better test the quality of the representations, we chose more comprehensive metrics to measure, the accuracy of the classification, the Precision score, the F1 score and the Recall. All of the baseline models are measured by the same metrics. 

\textbf{Data Preprocessing}:
For univariate time series data, we normalize the dataset using z-score following \cite{nips2019,yue2022ts2vec,zhou2021informer} to make sure that each of the datasets has zero mean and unit variance. For multivariate time series data, we normalize each variable independently by using z-score. 
Some of the datasets have missing values (\textit{NaN}), the corresponding position of missing value would be set to zero. 

\subsection{Implement Details}
For all datasets, a three-layer Fully Convolutional Network structure is used to encode and a projection head with two linear layers is applied. After the upstream encoder is reasonably trained, the learned representation will be passed to the SVM classifier with RBF kernel for classification. All experiments are conducted on a Linux machine with 4 NVIDIA GeForce RTX 3090 Ti GPUs, each with 24GB memory. CUDA version is 11.7 and Driver Version is 515.65.01. Our method RankSCL is implemented with Python 3.7.6 an Pytorch 1.12.0.
We train and evaluate our methods with the following hyper-parameters and configurations:
\textbf{Optimizer}: Adam optimizer \cite{kingma2014adam}
with learning rate and decay rates setting to 0.0001 and
0.0005. \textbf{SVM}: scikit-learn implementation \cite{pedregosa2011scikit} with penalty $C \in \left\{10^{i}|i \in \{-4, -3, ..., 4\} \cup \infty\right\} $ \cite{nips2019}. \textbf{Encoder architecture}: We choose a 3-layer Fully Convolutional Network to design the encoder. Specifically, the output dimension of the linear projection layer is set to 320, the same for the number of channels in the following projection head. \textbf{Classifier architecture}: a fully connected layer that maps the representations to the label is adopted. Number of augmentations: $k$ is searched in $\left\{0, 5, 10, 15\right\}$. 



\subsection{Univariate Time Series Results}
Table \ref{tab:table1} summarizes the experiment results from the UCR datasets. The Appendix contains the complete results in detail. RankSCL outperforms other baselines in terms of Precision score, F1 score and Recall value. It improves 0.1\% classification precision score, 0.4\% F1 score and 0.3\% Recall value. We noticed that these baseline methods just compare the accuracy values of the two datasets in their papers. To evaluate the performance of the different methods more reasonably and comprehensively, we also calculate and compare the F1 values and the Precision scores to make a comparison. The results of all the experiments are the results obtained by taking the mean value according to the 5 different seeds.

\begin{figure}

\begin{tikzpicture}[
  treatment line/.style={rounded corners=1.5pt, line cap=round, shorten >=1pt},
  treatment label/.style={font=\small},
  group line/.style={ultra thick},
]

\begin{axis}[
  clip={false},
  axis x line={center},
  axis y line={none},
  axis line style={-},
  xmin={1},
  ymax={0},
  scale only axis={true},
  width={\axisdefaultwidth},
  ticklabel style={anchor=south, yshift=1.3*\pgfkeysvalueof{/pgfplots/major tick length}, font=\small},
  every tick/.style={draw=black},
  major tick style={yshift=.5*\pgfkeysvalueof{/pgfplots/major tick length}},
  minor tick style={yshift=.5*\pgfkeysvalueof{/pgfplots/minor tick length}},
  title style={yshift=\baselineskip},
  xmax={8},
  ymin={-5.5},
  height={6\baselineskip},
  xtick={1,2,3,4,5,6,7,8},
  minor x tick num={1},
  x dir={reverse},
  title={critdd},
  scale=0.85
]

\draw[treatment line] ([yshift=-2pt] axis cs:2.4296875, 0) |- (axis cs:1.7630208333333335, -2.0)
  node[treatment label, anchor=west] {RankSCL};
\draw[treatment line] ([yshift=-2pt] axis cs:2.46484375, 0) |- (axis cs:1.7630208333333335, -3.0)
  node[treatment label, anchor=west] {TS2Vec};
\draw[treatment line] ([yshift=-2pt] axis cs:3.53125, 0) |- (axis cs:1.7630208333333335, -4.0)
  node[treatment label, anchor=west] {T-Loss};
\draw[treatment line] ([yshift=-2pt] axis cs:3.82421875, 0) |- (axis cs:1.7630208333333335, -5.0)
  node[treatment label, anchor=west] {InfoTS};
\draw[treatment line] ([yshift=-2pt] axis cs:5.2109375, 0) |- (axis cs:8.420572916666666, -5.0)
  node[treatment label, anchor=east] {TimesNet};
\draw[treatment line] ([yshift=-2pt] axis cs:5.27734375, 0) |- (axis cs:8.420572916666666, -4.0)
  node[treatment label, anchor=east] {DTW};
\draw[treatment line] ([yshift=-2pt] axis cs:5.5078125, 0) |- (axis cs:8.420572916666666, -3.0)
  node[treatment label, anchor=east] {TS-TCC};
\draw[treatment line] ([yshift=-2pt] axis cs:7.75390625, 0) |- (axis cs:8.420572916666666, -2.0)
  node[treatment label, anchor=east] {TNC};
\draw[group line] (axis cs:5.2109375, -2.0) -- (axis cs:5.5078125, -2.0);
\draw[group line] (axis cs:2.4296875, -1.3333333333333333) -- (axis cs:2.46484375, -1.3333333333333333);

\end{axis}
\end{tikzpicture}

\caption{Critical Difference (CD) diagram of  Univariate Time series classification task}
\label{cd_ucr}
\end{figure}
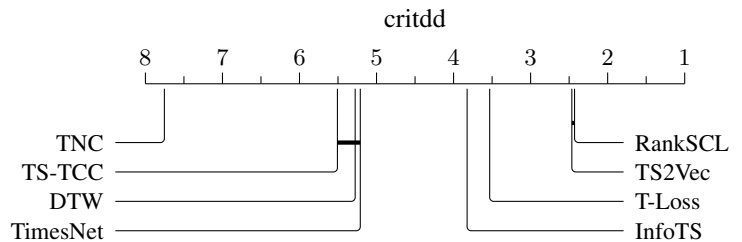

In particular, the RankSCL model achieves a similar accuracy score over the best baseline TS2Vec on all 128 UCR datasets. In terms of ACC, our method outperforms TS2Vec in 72 out of 128 UCR datasets, InfoTS \cite{InfoTS} in 96 datasets, and T-loss \cite{nips2019} in 88 datasets out of 125 datasets. Figure~\ref{cd_ucr} shows the Critical Difference diagram of different methods on 128 UCR datasets. This result indicates that RankSCL has similar results with the TS2Vec \cite{ts2vec} model in terms of Accuracy scores, while significantly outperforming the other methods. More detailed results of 128 UCR datasets are shown in the Appendix Table\ref{tab4}.

\begin{table*}[h!tb]
  \centering
 \small
  
    \begin{tabular}{lcccc}
    \toprule
     & Avg. Accuracy $\uparrow$ & Avg. Precision $\uparrow$ & Avg. F1 $\uparrow$ & Avg. Recall $\uparrow$\\
     \hline
     \textbf{RankSCL} & \textbf{0.797} & \textbf{0.834}&\textbf{0.793} & \textbf{0.756} \\
     \hline
    $\textbf{w}\textbf{/}\textbf{o}$ Data Augmentation &0.636 &0.663 &0.62 &0.582\\
    $\textbf{w}\textbf{/}$ Data Augmentation (Raw Data) &0.519 &0.550 &0.500 &0.458 \\
    $\textbf{w}\textbf{/}\textbf{o}$ FCN (Resnet Backbone) &0.318 &0.319 &0.293 &0.271\\
    $\textbf{w}\textbf{/}\textbf{o}$ Rank Loss (CE) &0.636 &0.663 &0.620 &0.582\\
     $\textbf{w}\textbf{/}\textbf{o}$ Rank Loss (CTL) &0.676 &0.710 &0.663 &0.622\\
    $\textbf{w}\textbf{/}\textbf{o}$ Rank Loss (TL) & 0.506&0.532 &0.509 &0.488\\

     \hline
    
 \end{tabular}
  
   \caption{Ablation studies on PigCVP dataset}
   \label{tab:table3}%
\end{table*}%

\begin{table*}[htbp]
\centering
\small
\begin{tabular}{ccccc}
\toprule
 Number & Avg. Accuracy $\uparrow$ & Avg. Precision $\uparrow$ & Avg. F1 $\uparrow$ & Avg. Recall $\uparrow$\\
     \hline
   
    0 &0.636 &0.663 &0.62 &0.582\\
    \textbf{5} & \textbf{0.797} & \textbf{0.834}&\textbf{0.793} & \textbf{0.756} \\
    10 & 0.517 & 0.544 & 0.503& 0.468 \\
    15 & 0.520 & 0.560 & 0.504 & 0.458\\
    \hline
\end{tabular}

\caption{Data Augmentation Analysis on PigCVP dataset}\vspace{-5mm}
\label{tab:table4}%
\end{table*}

\subsection{Multivariate Time Series Results}
Table \ref{tab:table1} also shows the comparison results on UEA datasets for the multivariate time series classification. The proposed RankSCL significantly outperforms all baseline methods in terms of accuracy, precision score, F1 score, and Recall value. In particular,  RankSCL outperforms the best baseline model TS2Vec 
 \cite{ts2vec} by 2.0\% in terms of Accuracy, by 3.2\% in precision score, by 3.3\% in F1 score and 3.5\% by Recall value. The results indicate that the proposed augmentation techniques coupled with the rank loss function are more effective on multivariate time series data compared to univariate time series data. More detailed results of 30 UEA datasets are shown in the Appendix Table\ref{tab5}.

\begin{figure}
\begin{tikzpicture}[
  treatment line/.style={rounded corners=1.5pt, line cap=round, shorten >=1pt},
  treatment label/.style={font=\small},
  group line/.style={ultra thick},
]

\begin{axis}[
  clip={false},
  axis x line={center},
  axis y line={none},
  axis line style={-},
  xmin={1},
  ymax={0},
  scale only axis={true},
  width={\axisdefaultwidth},
  ticklabel style={anchor=south, yshift=1.3*\pgfkeysvalueof{/pgfplots/major tick length}, font=\small},
  every tick/.style={draw=black},
  major tick style={yshift=.5*\pgfkeysvalueof{/pgfplots/major tick length}},
  minor tick style={yshift=.5*\pgfkeysvalueof{/pgfplots/minor tick length}},
  title style={yshift=\baselineskip},
  xmax={8},
  ymin={-5.5},
  height={6\baselineskip},
  xtick={1,2,3,4,5,6,7,8},
  minor x tick num={1},
  x dir={reverse},
  title={critdd},
  scale=0.85
]

\draw[treatment line] ([yshift=-2pt] axis cs:2.433333333333333, 0) |- (axis cs:1.7666666666666666, -2.0)
  node[treatment label, anchor=west] {RankSCL};
\draw[treatment line] ([yshift=-2pt] axis cs:3.65, 0) |- (axis cs:1.7666666666666666, -3.0)
  node[treatment label, anchor=west] {TS2Vec};
\draw[treatment line] ([yshift=-2pt] axis cs:3.9166666666666665, 0) |- (axis cs:1.7666666666666666, -4.0)
  node[treatment label, anchor=west] {InfoTS};
\draw[treatment line] ([yshift=-2pt] axis cs:4.233333333333333, 0) |- (axis cs:1.7666666666666666, -5.0)
  node[treatment label, anchor=west] {TimesNet};
\draw[treatment line] ([yshift=-2pt] axis cs:4.483333333333333, 0) |- (axis cs:7.6000000000000005, -5.0)
  node[treatment label, anchor=east] {DTW};
\draw[treatment line] ([yshift=-2pt] axis cs:4.8, 0) |- (axis cs:7.6000000000000005, -4.0)
  node[treatment label, anchor=east] {T-Loss};
\draw[treatment line] ([yshift=-2pt] axis cs:5.55, 0) |- (axis cs:7.6000000000000005, -3.0)
  node[treatment label, anchor=east] {TS-TCC};
\draw[treatment line] ([yshift=-2pt] axis cs:6.933333333333334, 0) |- (axis cs:7.6000000000000005, -2.0)
  node[treatment label, anchor=east] {TNC};
\draw[group line] (axis cs:3.9166666666666665, -2.0) -- (axis cs:5.55, -2.0);
\draw[group line] (axis cs:3.65, -2.2) -- (axis cs:4.8, -2.2);
\draw[group line] (axis cs:2.433333333333333, -1.3333333333333333) -- (axis cs:4.483333333333333, -1.3333333333333333);

\end{axis}
\end{tikzpicture}

\caption{Critical Difference (CD) diagram of  Multivariate Time series classification task}\vspace{-3mm}
\label{cduea}
\end{figure}
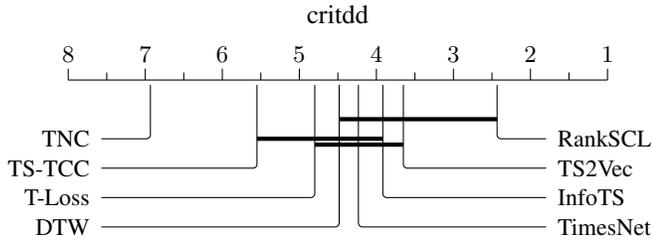

\begin{figure*}[!htb]
    \centering
    \subfigure[SyntheticControl]{
        \includegraphics[width=2.15in]{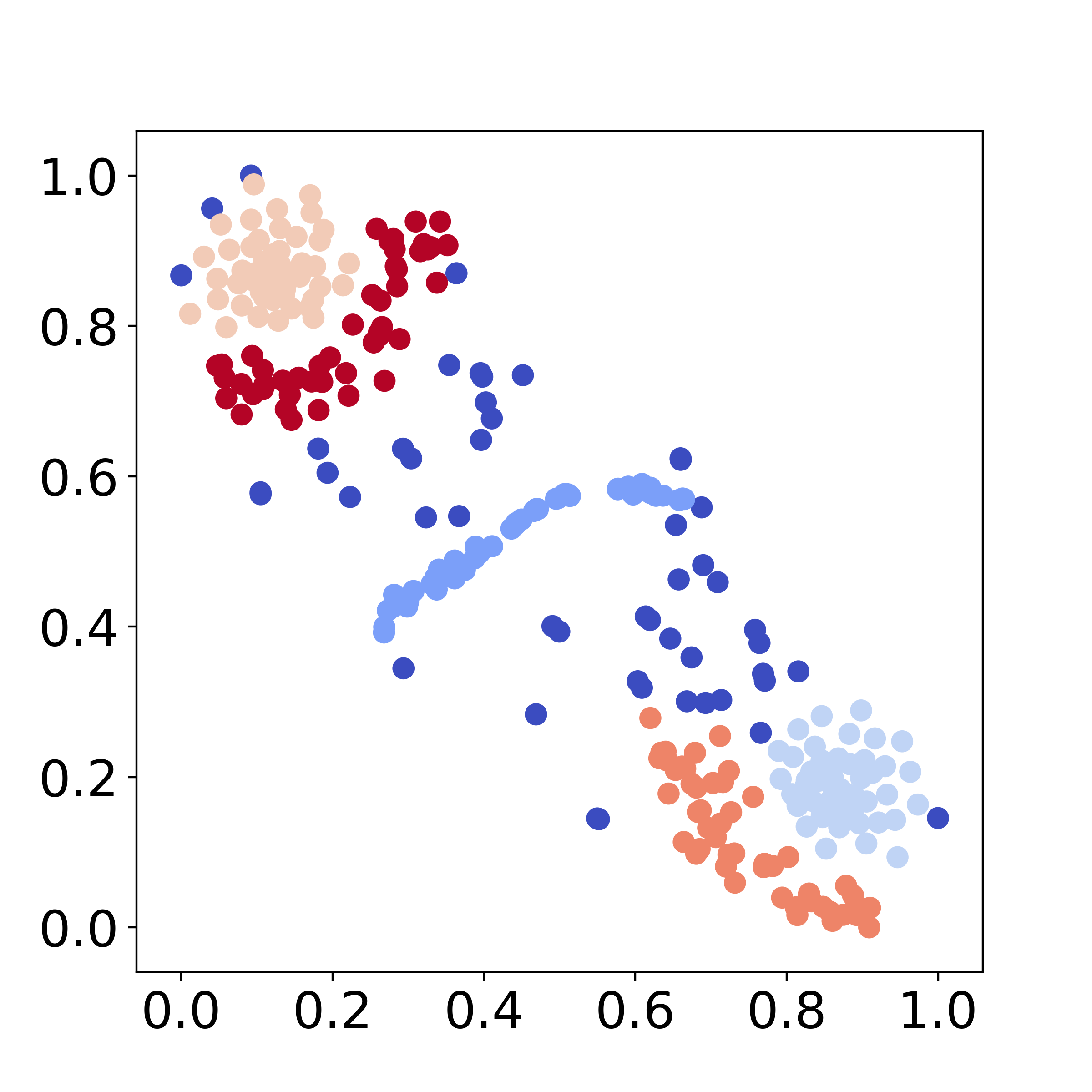}
        \label{label_for_cross_ref_1}
    }
    \subfigure[TS2Vec]{
    	\includegraphics[width=2.15in]{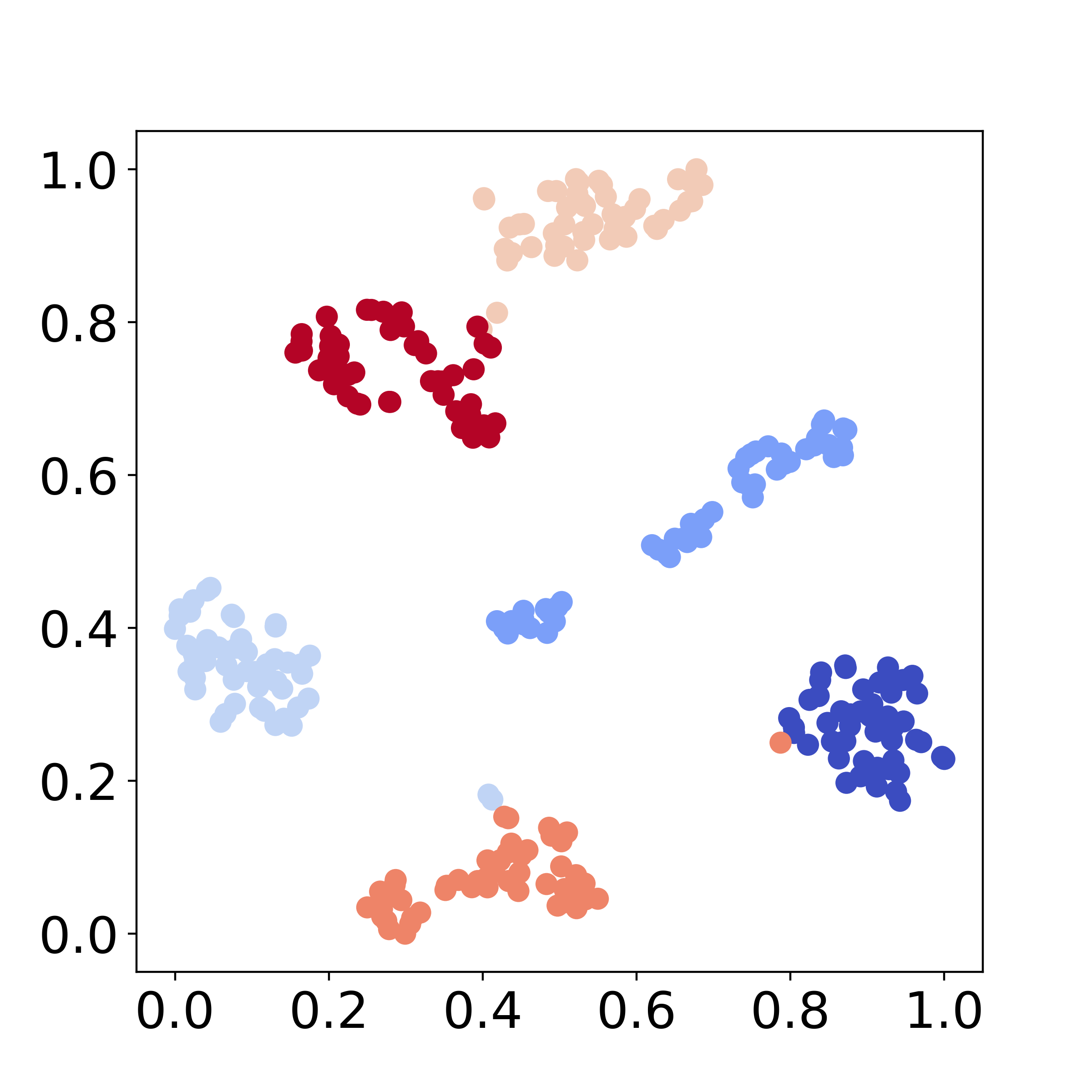}
        \label{label_for_cross_ref_3}
    }
    \subfigure[RankSCL]{
	\includegraphics[width=2.15in]{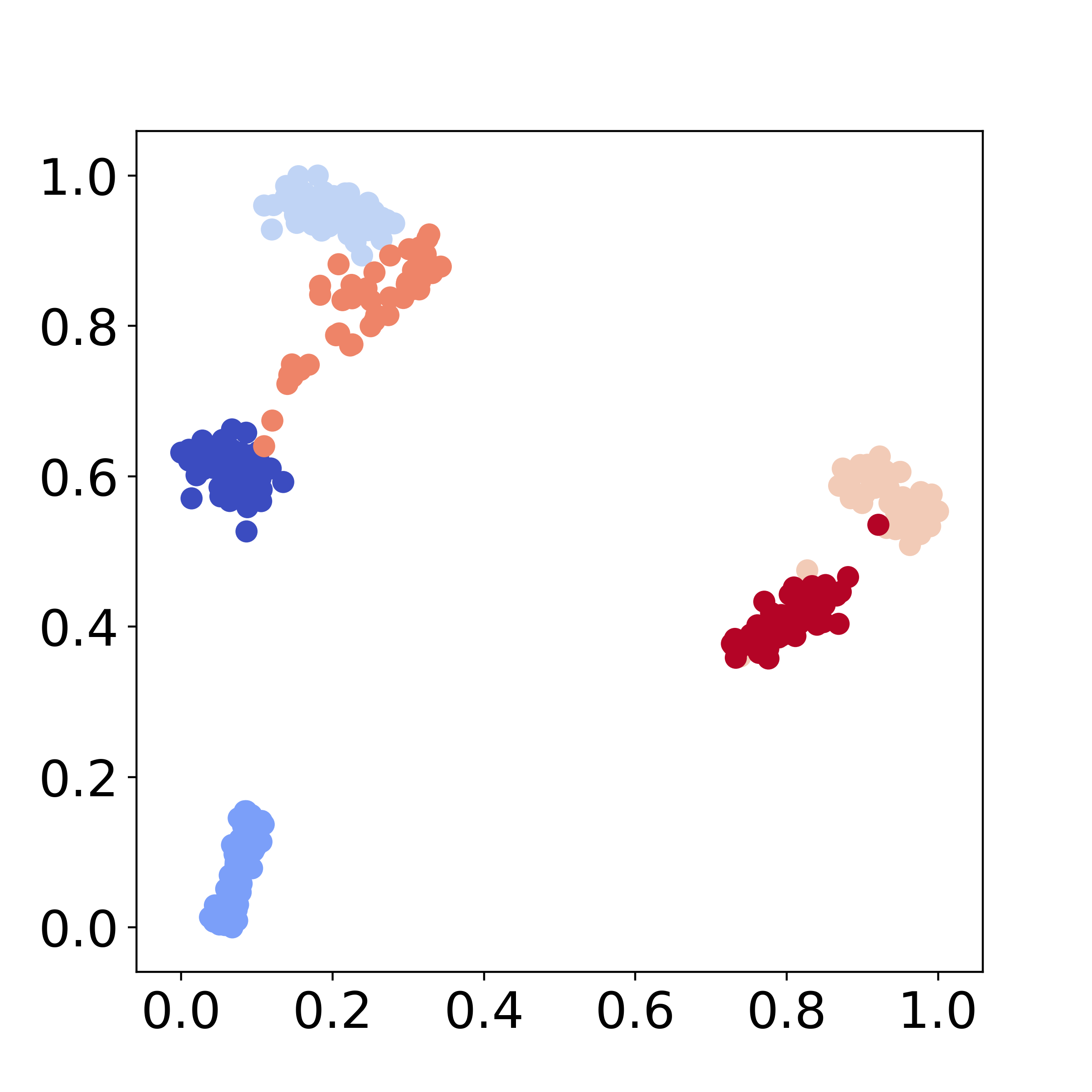}
        \label{label_for_cross_ref_4}
    }
    \caption{The t-SNE visualization of the learned representations of SyntheticControl dataset (with 6 classes). (Best viewed in color)}
    \label{tsne}\vspace{-1mm}
\end{figure*}

\subsection{Ablation Study}
A full comparison of different variants of RankSCL model is evaluated on PigCVP dataset to justify the effectiveness of our method and the results are shown in Table \ref{tab:table3}. To assess the effectiveness of the data augmentation, we evaluate different numbers of augmented positive samples. The average accuracy without augmented positive samples significantly drops 16.1\%, and this is even more obvious in terms of other metrics. After the number of augmented samples exceeds 5, there is a significant drop in performance, proving that the number of augmented positive samples does not linearly increase with performance. We also evaluate the experiment of data augmentation on raw time series data (\textbf{w}\textbf{/}\textbf{o} Data Augmentation (Raw Data)). In Table \ref{tab:table3}, we can observe it is not as effective as augmenting the samples in the embedding space (\textbf{w}\textbf{/}\textbf{o} Data Augmentation), which can demonstrate the effectiveness of our proposed method. Different encoder architectures are employed for comparison in Table \ref{tab:table3}. We observe that FCN architecture is more effective than other variations (\textbf{w}\textbf{/}\textbf{o} FCN (Resnet Backbone)) on the PigCVP dataset. We also make comparisons of different loss functions, including cross entropy(CE), contrastive loss(CTL), and triplet loss(TL). The results suggest that Rank based arctan loss function is better than other alternatives.

\subsection{Qualitative Evaluation}
Based on t-SNE \cite{t-sne}, we project SyntheticControl dataset (with 6 classes) from raw space to the two-dimensional space based on t-SNE \cite{t-sne} (Figure~\ref{tsne}(a)). Similarly, for TS2Vec and RankSCL, we also project their learned embeddings to the two-dimensional space with t-SNE and compare their visualization results accordingly. We can observe for both TS2Vec and RankSCL, different classes are well separated and clustered together. Moreover, TS2Vec has 7 clusters while RankSCL has 6 clusters which is consistent with the number of classes 6. This demonstrates that our proposed RankSCL can preserve more clear boundaries among the different classes.

\section{Conclusion}
In this paper, we propose a novel supervised contrastive learning framework termed RankSCL which learns optimal time series representations for classfication task. A new data augmentation method is proposed to generate more targeted positive samples in the embedding space, which enriches the intra-class information. Moreover, a certain mining rule is applied to capture valid triplet pairs to reduce the computational complexity. We further propose a novel rank loss function that sorts different positive samples to learn optimal representations. The evaluation results prove the effectiveness of these strategies and certify that with proper data augmentation methodology, enough valid triplet pairs, and abundant intra-class information, the representations learned by RankSCL are highly qualified to be applied to more future tasks and other modalities of data.

{
\bibliographystyle{IEEEtran}
\bibliography{IEEEexample}
}

\vspace{-5mm}
\onecolumn
{
\tiny

\setlength{\tabcolsep}{4pt}

\begin{longtable}{lcccccccccccccc}

    \hline
Dataset & \multicolumn{2}{c}{RankSCL} & \multicolumn{2}{c}{InfoTS} & \multicolumn{2}{c}{TS2Vec} & \multicolumn{2}{c}{T-Loss} & \multicolumn{2}{c}{DTW} & \multicolumn{2}{c}{TS-TCC} & \multicolumn{2}{c}{TNC} \\
\cmidrule(r){2-3}  \cmidrule(r){4-5} \cmidrule(r){6-7}  \cmidrule(r){8-9}  \cmidrule(r){10-11}  \cmidrule(r){12-13}  \cmidrule(r){14-15} 
& ACC & Prec. & ACC & Prec. & ACC & Prec. & ACC & Prec. & ACC & Prec. & ACC & Prec. & ACC & Prec. \\ \hline
\endfirsthead

\multicolumn{15}{c} 
{{\bfseries Table \thetable\ continued from previous page}} \\
\hline
Dataset & \multicolumn{2}{c}{RankSCL} & \multicolumn{2}{c}{InfoTS} & \multicolumn{2}{c}{TS2Vec} & \multicolumn{2}{c}{T-Loss} & \multicolumn{2}{c}{DTW} & \multicolumn{2}{c}{TS-TCC} & \multicolumn{2}{c}{TNC} \\ \cmidrule(r){2-3}  \cmidrule(r){4-5} \cmidrule(r){6-7}  \cmidrule(r){8-9}  \cmidrule(r){10-11}  \cmidrule(r){12-13}  \cmidrule(r){14-15} 
& ACC & Prec. & ACC & Prec. & ACC & Prec. & ACC & Prec. & ACC & Prec. & ACC & Prec. & ACC & Prec. \\ \hline
\endhead

\hline 
\multicolumn{15}{r}{{Continued on next page}} \\ \hline
\endfoot

\endlastfoot
    
       Adiac & \textbf{0.822} & \textbf{0.821} & 0.753 & 0.768 & 0.765 & 0.772 & 0.645 & 0.642 & 0.573 & 0.624 & 0.241 & 0.148 & 0.068 & 0.029 \\
ACSF1 & \textbf{0.896} & \textbf{0.906} & 0.772 & 0.797 & 0.870 & 0.883 & \textbf{0.896} & 0.896 & 0.590 & 0.630 & 0.438 & 0.367 & 0.166 & 0.065 \\
ArrowHead & \textbf{0.848} & \textbf{0.853} & 0.824 & 0.832 & 0.834 & 0.833 & 0.804 & 0.813 & 0.709 & 0.705 & 0.694 & 0.554 & 0.339 & 0.113 \\
Beef & \textbf{0.747} & \textbf{0.785} & 0.738 & 0.773 & 0.720 & 0.781 & 0.660 & 0.660 & 0.567 & 0.639 & 0.654 & 0.471 & 0.240 & 0.172 \\
BeetleFly & \textbf{0.940} & \textbf{0.947} & 0.826 & 0.871 & 0.890 & 0.916 & 0.880 & 0.560 & 0.700 & 0.738 & 0.561 & 0.673 & 0.580 & 0.540 \\
BirdChicken & \textbf{0.900} & \textbf{0.917} & 0.724 & 0.732 & 0.870 & 0.887 & 0.810 & 0.850 & 0.600 & 0.600 & 0.717 & 0.682 & 0.540 & 0.402 \\
Car & \textbf{0.847} & \textbf{0.853} & 0.770 & 0.765 & 0.730 & 0.744 & 0.713 & 0.718 & 0.517 & 0.546 & 0.728 & 0.667 & 0.217 & 0.062 \\
CBF & \textbf{0.997} & \textbf{0.997} & 0.991 & 0.992 & 0.996 & 0.996 & 0.987 & 0.987 & \textbf{0.997} & \textbf{0.997} & 0.822 & 0.830 & 0.384 & 0.352 \\
ChlorineConcentration & 0.817 & 0.799 & 0.779 & 0.750 & \textbf{0.823} & \textbf{0.804} & 0.724 & 0.682 & 0.548 & 0.484 & 0.565 & 0.306 & 0.549 & 0.447 \\
CinCECGTorso & 0.720 & 0.741 & 0.803 & 0.816 & 0.798 & 0.814 & 0.682 & 0.682 & 0.498 & 0.534 & \textbf{0.858} & \textbf{0.834} & 0.272 & 0.160 \\
Coffee & \textbf{1.000} & \textbf{1.000} & \textbf{1.000} & \textbf{1.000} & \textbf{1.000} & \textbf{1.000} & \textbf{1.000} & \textbf{1.000} & 0.929 & 0.941 & 0.881 & 0.825 & 0.507 & 0.254 \\
Computers & \textbf{0.763} & \textbf{0.764} & 0.655 & 0.669 & 0.652 & 0.653 & 0.669 & 0.577 & 0.684 & 0.687 & 0.562 & 0.513 & 0.552 & 0.506 \\
CricketX & \textbf{0.792} & 0.799 & 0.699 & 0.715 & 0.789 & \textbf{0.809} & 0.774 & 0.778 & 0.744 & 0.767 & 0.662 & 0.573 & 0.138 & 0.081 \\
CricketY & \textbf{0.789} & \textbf{0.793} & 0.693 & 0.711 & 0.762 & 0.775 & 0.733 & 0.733 & 0.685 & 0.718 & 0.637 & 0.534 & 0.175 & 0.117 \\
CricketZ & \textbf{0.823} & \textbf{0.823} & 0.725 & 0.731 & 0.800 & 0.800 & 0.760 & 0.755 & 0.754 & 0.763 & 0.680 & 0.585 & 0.112 & 0.094 \\
DiatomSizeReduction & 0.924 & 0.946 & \textbf{0.982} & \textbf{0.979} & 0.950 & 0.970 & 0.976 & 0.948 & 0.709 & 0.616 & 0.665 & 0.546 & 0.311 & 0.101 \\
DistalPhalanxOutlineCorrect & \textbf{0.766} & 0.760 & 0.759 & 0.762 & 0.764 & 0.767 & \textbf{0.766} & \textbf{0.781} & 0.732 & 0.754 & 0.649 & 0.552 & 0.595 & 0.414 \\
DistalPhalanxOutlineAgeGroup & 0.741 & 0.730 & 0.737 & \textbf{0.777} & 0.716 & 0.745 & 0.727 & 0.696 & \textbf{0.748} & 0.754 & 0.707 & 0.535 & 0.643 & 0.436 \\
DistalPhalanxTW & \textbf{0.692} & \textbf{0.543} & 0.670 & 0.497 & 0.682 & 0.507 & 0.668 & 0.515 & 0.662 & 0.498 & 0.689 & 0.490 & 0.579 & 0.205 \\
Earthquakes & 0.732 & 0.601 & 0.749 & 0.475 & 0.748 & 0.374 & 0.748 & 0.500 & 0.748 & 0.628 & 0.736 & 0.615 & \textbf{0.754} & \textbf{0.690} \\
ECG200 & \textbf{0.920} & \textbf{0.917} & 0.864 & 0.857 & 0.908 & 0.891 & 0.862 & 0.873 & 0.790 & 0.781 & 0.841 & 0.805 & 0.684 & 0.633 \\
ECG5000 & \textbf{0.942} & 0.680 & 0.939 & 0.733 & 0.938 & 0.699 & 0.935 & 0.514 & 0.938 & 0.655 & 0.941 & \textbf{0.874} & 0.873 & 0.391 \\
ECGFiveDays & 0.997 & 0.997 & 0.987 & 0.987 & \textbf{1.000} & \textbf{1.000} & 0.999 & 0.975 & 0.598 & 0.683 & 0.834 & 0.844 & 0.547 & 0.427 \\
ElectricDevices & 0.691 & 0.652 & 0.677 & 0.648 & 0.710 & \textbf{0.678} & \textbf{0.714} & 0.620 & 0.610 & 0.578 & 0.672 & 0.554 & 0.443 & 0.342 \\
FaceAll & 0.775 & 0.812 & 0.753 & 0.792 & 0.783 & 0.814 & 0.758 & \textbf{0.854} & \textbf{0.808} & 0.811 & 0.765 & 0.703 & 0.239 & 0.257 \\
FaceFour & 0.832 & 0.833 & 0.764 & 0.777 & 0.886 & 0.906 & 0.805 & 0.827 & 0.682 & 0.746 & \textbf{0.932} & \textbf{0.913} & 0.359 & 0.260 \\
FacesUCR & 0.914 & 0.919 & 0.848 & 0.851 & \textbf{0.932} & \textbf{0.934} & 0.875 & 0.826 & 0.886 & 0.880 & 0.893 & 0.837 & 0.295 & 0.158 \\
FiftyWords & 0.727 & 0.653 & 0.758 & 0.678 & \textbf{0.775} & \textbf{0.734} & 0.732 & 0.578 & 0.664 & 0.561 & 0.710 & 0.564 & 0.148 & 0.016 \\
Fish & \textbf{0.941} & \textbf{0.954} & 0.829 & 0.845 & 0.930 & 0.934 & 0.844 & 0.847 & 0.794 & 0.793 & 0.746 & 0.693 & 0.155 & 0.042 \\
FordA & \textbf{0.956} & \textbf{0.956} & 0.878 & 0.879 & 0.940 & 0.940 & 0.931 & 0.910 & 0.594 & 0.595 & 0.912 & 0.906 & 0.524 & 0.522 \\
FordB & \textbf{0.828} & \textbf{0.829} & 0.730 & 0.730 & 0.802 & 0.806 & 0.794 & 0.773 & 0.654 & 0.661 & 0.763 & 0.757 & 0.506 & 0.505 \\
GunPoint & \textbf{0.995} & \textbf{0.995} & 0.975 & 0.975 & 0.983 & 0.983 & 0.976 & 0.943 & 0.827 & 0.829 & 0.879 & 0.884 & 0.513 & 0.506 \\
Ham & 0.714 & 0.715 & 0.695 & 0.697 & 0.711 & 0.711 & 0.659 & 0.615 & 0.590 & 0.590 & \textbf{0.754} & \textbf{0.726} & 0.520 & 0.443 \\
HandOutlines & 0.913 & 0.906 & 0.924 & 0.922 & \textbf{0.928} & \textbf{0.931} & 0.909 & 0.896 & 0.889 & 0.900 & 0.663 & 0.671 & 0.641 & 0.370 \\
Haptics & \textbf{0.539} & \textbf{0.544} & 0.499 & 0.508 & 0.526 & 0.534 & 0.451 & 0.446 & 0.406 & 0.437 & 0.367 & 0.297 & 0.213 & 0.176 \\
Herring & 0.613 & \textbf{0.614} & \textbf{0.621} & 0.564 & 0.600 & 0.514 & 0.556 & 0.516 & 0.547 & 0.544 & 0.541 & 0.532 & 0.594 & 0.297 \\
InlineSkate & 0.371 & 0.384 & 0.350 & 0.376 & \textbf{0.389} & \textbf{0.426} & 0.357 & 0.357 & 0.356 & 0.388 & 0.286 & 0.227 & 0.169 & 0.049 \\
InsectWingbeatSound & 0.615 & 0.614 & 0.630 & \textbf{0.634} & 0.626 & 0.628 & 0.564 & 0.565 & 0.362 & 0.394 & \textbf{0.662} & 0.565 & 0.186 & 0.097 \\
ItalyPowerDemand & \textbf{0.960} & \textbf{0.960} & 0.957 & 0.957 & 0.958 & 0.959 & 0.946 & 0.901 & 0.946 & 0.946 & 0.940 & 0.935 & 0.746 & 0.783 \\
LargeKitchenAppliances & \textbf{0.912} & \textbf{0.915} & 0.771 & 0.773 & 0.865 & 0.871 & 0.820 & 0.820 & 0.800 & 0.805 & 0.585 & 0.445 & 0.386 & 0.424 \\
Lightning2 & 0.830 & 0.833 & 0.743 & 0.781 & \textbf{0.852} & \textbf{0.856} & \textbf{0.852} & 0.799 & 0.820 & 0.855 & 0.709 & 0.712 & 0.577 & 0.454 \\
Lightning7 & 0.786 & 0.820 & 0.762 & 0.811 & \textbf{0.844} & \textbf{0.880} & 0.784 & 0.760 & 0.726 & 0.852 & 0.768 & 0.689 & 0.260 & 0.159 \\
Mallat & \textbf{0.962} & \textbf{0.966} & 0.879 & 0.897 & 0.906 & 0.917 & 0.959 & 0.959 & 0.928 & 0.940 & 0.820 & 0.745 & 0.156 & 0.074 \\
Meat & \textbf{0.967} & \textbf{0.970} & 0.920 & 0.930 & 0.903 & 0.923 & 0.910 & 0.910 & 0.933 & 0.944 & 0.313 & 0.281 & 0.333 & 0.111 \\
MedicalImages & \textbf{0.793} & 0.786 & 0.765 & 0.754 & 0.788 & \textbf{0.808} & 0.762 & 0.714 & 0.709 & 0.694 & 0.690 & 0.501 & 0.513 & 0.118 \\
MiddlePhalanxOutlineCorrect & \textbf{0.846} & \textbf{0.846} & 0.801 & 0.798 & 0.789 & 0.786 & 0.777 & 0.756 & 0.732 & 0.730 & 0.619 & 0.551 & 0.570 & 0.285 \\
MiddlePhalanxOutlineAgeGroup & 0.445 & 0.447 & \textbf{0.651} & \textbf{0.874} & 0.641 & 0.813 & 0.644 & 0.472 & 0.558 & 0.474 & 0.636 & 0.413 & 0.554 & 0.330 \\
MiddlePhalanxTW & 0.579 & \textbf{0.458} & \textbf{0.589} & 0.441 & 0.586 & 0.425 & 0.574 & 0.417 & 0.552 & 0.350 & 0.576 & 0.308 & 0.500 & 0.178 \\
MoteStrain & \textbf{0.889} & \textbf{0.888} & 0.861 & 0.868 & 0.871 & 0.870 & 0.857 & 0.742 & 0.814 & 0.816 & 0.844 & 0.847 & 0.737 & 0.756 \\
NonInvasiveFetalECGThorax1 & \textbf{0.951} & \textbf{0.949} & 0.934 & 0.932 & 0.925 & 0.923 & 0.890 & 0.889 & 0.786 & 0.808 & 0.863 & 0.784 & 0.167 & 0.108 \\
NonInvasiveFetalECGThorax2 & 0.893 & 0.891 & \textbf{0.937} & 0.935 & \textbf{0.937} & \textbf{0.936} & 0.925 & 0.923 & 0.863 & 0.865 & 0.848 & 0.754 & 0.138 & 0.080 \\
OliveOil & 0.840 & 0.851 & \textbf{0.868} & \textbf{0.909} & 0.840 & 0.896 & 0.833 & 0.777 & 0.767 & 0.694 & 0.407 & 0.303 & 0.400 & 0.100 \\
OSULeaf & \textbf{0.926} & \textbf{0.927} & 0.659 & 0.658 & 0.831 & 0.843 & 0.746 & 0.734 & 0.545 & 0.498 & 0.498 & 0.433 & 0.234 & 0.174 \\
PhalangesOutlinesCorrect & \textbf{0.829} & \textbf{0.826} & 0.795 & 0.813 & 0.809 & 0.810 & 0.799 & 0.786 & 0.755 & 0.784 & 0.656 & 0.500 & 0.615 & 0.508 \\
Phoneme & 0.277 & \textbf{0.169} & 0.198 & 0.089 & \textbf{0.307} & 0.157 & 0.267 & 0.115 & 0.294 & 0.144 & 0.182 & 0.135 & 0.111 & 0.017 \\
Plane & 0.996 & 0.996 & 0.996 & 0.997 & 0.986 & 0.988 & 0.988 & 0.988 & \textbf{1.000} & \textbf{1.000} & 0.964 & 0.939 & 0.263 & 0.189 \\
ProximalPhalanxOutlineCorrect & \textbf{0.893} & 0.888 & 0.874 & \textbf{0.890} & 0.882 & 0.883 & 0.867 & 0.838 & 0.832 & 0.849 & 0.711 & 0.583 & 0.692 & 0.537 \\
ProximalPhalanxOutlineAgeGroup & 0.843 & 0.742 & \textbf{0.859} & \textbf{0.769} & 0.845 & 0.735 & 0.848 & 0.743 & 0.824 & 0.709 & 0.645 & 0.561 & 0.855 & 0.637 \\
ProximalPhalanxTW & 0.776 & 0.547 & \textbf{0.819} & \textbf{0.603} & 0.793 & 0.570 & 0.801 & 0.550 & 0.761 & 0.499 & 0.643 & 0.394 & 0.674 & 0.235 \\
RefrigerationDevices & 0.554 & 0.575 & 0.566 & 0.596 & \textbf{0.596} & \textbf{0.627} & 0.524 & 0.525 & 0.475 & 0.489 & 0.452 & 0.382 & 0.346 & 0.332 \\
ScreenType & \textbf{0.583} & \textbf{0.589} & 0.447 & 0.447 & 0.411 & 0.389 & 0.421 & 0.424 & 0.427 & 0.440 & 0.381 & 0.374 & 0.337 & 0.320 \\
ShapeletSim & 0.949 & 0.949 & 0.556 & 0.557 & \textbf{0.990} & \textbf{0.990} & 0.759 & 0.532 & 0.639 & 0.640 & 0.524 & 0.481 & 0.509 & 0.508 \\
ShapesAll & \textbf{0.908} & \textbf{0.921} & 0.789 & 0.799 & 0.896 & 0.909 & 0.851 & 0.852 & 0.708 & 0.687 & 0.718 & 0.563 & 0.072 & 0.018 \\
SmallKitchenAppliances & \textbf{0.813} & \textbf{0.822} & 0.738 & 0.735 & 0.727 & 0.724 & 0.686 & 0.686 & 0.688 & 0.684 & 0.610 & 0.441 & 0.513 & 0.435 \\
SonyAIBORobotSurface1 & 0.831 & 0.858 & 0.864 & 0.877 & \textbf{0.885} & 0.892 & 0.835 & \textbf{0.973} & 0.617 & 0.764 & 0.557 & 0.575 & 0.448 & 0.364 \\
SonyAIBORobotSurface2 & \textbf{0.935} & \textbf{0.929} & 0.860 & 0.852 & 0.870 & 0.862 & 0.920 & 0.813 & 0.803 & 0.801 & 0.813 & 0.800 & 0.681 & 0.663 \\
StarLightCurves & \textbf{0.974} & \textbf{0.962} & 0.962 & 0.943 & 0.968 & 0.952 & 0.961 & 0.947 & 0.914 & 0.871 & 0.939 & 0.901 & 0.818 & 0.614 \\
Strawberry & \textbf{0.972} & \textbf{0.966} & 0.960 & 0.952 & 0.963 & 0.955 & 0.945 & 0.931 & 0.919 & 0.906 & 0.897 & 0.820 & 0.644 & 0.336 \\
SwedishLeaf & \textbf{0.955} & \textbf{0.956} & 0.911 & 0.913 & 0.939 & 0.940 & 0.926 & 0.927 & 0.779 & 0.787 & 0.884 & 0.813 & 0.231 & 0.195 \\
Symbols & 0.859 & 0.875 & 0.955 & 0.956 & \textbf{0.971} & \textbf{0.972} & 0.958 & 0.958 & 0.929 & 0.943 & 0.834 & 0.775 & 0.262 & 0.189 \\
SyntheticControl & \textbf{0.995} & \textbf{0.995} & \textbf{0.995} & \textbf{0.995} & \textbf{0.995} & \textbf{0.995} & 0.984 & 0.984 & 0.983 & 0.984 & 0.990 & 0.969 & 0.695 & 0.707 \\
ToeSegmentation1 & \textbf{0.961} & \textbf{0.962} & 0.808 & 0.818 & 0.914 & 0.918 & 0.943 & 0.790 & 0.741 & 0.787 & 0.576 & 0.514 & 0.485 & 0.370 \\
ToeSegmentation2 & 0.780 & 0.714 & 0.816 & 0.718 & \textbf{0.868} & \textbf{0.790} & 0.852 & 0.762 & 0.815 & 0.701 & 0.574 & 0.551 & 0.527 & 0.422 \\
Trace & \textbf{1.000} & \textbf{1.000} & \textbf{1.000} & \textbf{1.000} & 0.998 & 0.998 & \textbf{1.000} & \textbf{1.000} & 0.963 & \textbf{1.000} & 0.702 & 0.616 & 0.350 & 0.248 \\
TwoLeadECG & \textbf{0.996} & \textbf{0.996} & 0.924 & 0.932 & 0.979 & 0.980 & 0.991 & 0.922 & 0.852 & 0.883 & 0.724 & 0.730 & 0.537 & 0.577 \\
TwoPatterns & \textbf{1.000} & \textbf{1.000} & 0.999 & 0.999 & \textbf{1.000} & \textbf{1.000} & \textbf{1.000} & \textbf{1.000} & \textbf{1.000} & \textbf{1.000} & 0.999 & 0.999 & 0.402 & 0.372 \\
UWaveGestureLibraryX & \textbf{0.804} & \textbf{0.801} & \textbf{0.804} & 0.796 & \textbf{0.804} & 0.799 & 0.802 & \textbf{0.801} & 0.736 & 0.708 & 0.767 & 0.703 & 0.289 & 0.179 \\
UWaveGestureLibraryY & \textbf{0.724} & 0.724 & 0.722 & 0.720 & 0.723 & 0.722 & \textbf{0.724} & \textbf{0.727} & 0.632 & 0.626 & 0.640 & 0.551 & 0.405 & 0.322 \\
UWaveGestureLibraryZ & 0.756 & 0.749 & 0.741 & 0.732 & \textbf{0.759} & \textbf{0.752} & 0.748 & 0.751 & 0.671 & 0.659 & 0.683 & 0.598 & 0.336 & 0.228 \\
UWaveGestureLibraryAll & 0.836 & 0.838 & \textbf{0.954} & \textbf{0.954} & 0.934 & 0.934 & 0.908 & 0.909 & 0.179 & 0.181 & 0.939 & 0.914 & 0.316 & 0.269 \\
Wafer & \textbf{0.996} & 0.990 & 0.995 & \textbf{0.991} & \textbf{0.996} & \textbf{0.991} & 0.992 & 0.987 & 0.975 & 0.975 & 0.994 & 0.985 & 0.895 & 0.716 \\
Wine & 0.789 & 0.790 & 0.813 & \textbf{0.820} & 0.763 & 0.713 & \textbf{0.830} & 0.815 & 0.574 & 0.575 & 0.496 & 0.473 & 0.500 & 0.300 \\
WordSynonyms & 0.663 & 0.616 & 0.654 & 0.581 & \textbf{0.690} & \textbf{0.659} & 0.674 & 0.500 & 0.596 & 0.575 & 0.608 & 0.430 & 0.237 & 0.028 \\
Worms & \textbf{0.821} & \textbf{0.834} & 0.646 & 0.702 & 0.719 & 0.751 & 0.709 & 0.612 & 0.403 & 0.381 & 0.537 & 0.417 & 0.424 & 0.131 \\
WormsTwoClass & 0.753 & \textbf{0.782} & 0.693 & 0.692 & 0.784 & 0.781 & \textbf{0.808} & 0.726 & 0.532 & 0.537 & 0.583 & 0.532 & 0.561 & 0.434 \\
Yoga & \textbf{0.895} & \textbf{0.895} & 0.829 & 0.829 & 0.881 & 0.882 & 0.871 & 0.836 & 0.785 & 0.785 & 0.733 & 0.732 & 0.516 & 0.554 \\
AllGestureWiimoteX & 0.752 & 0.761 & 0.109 & 0.029 & \textbf{0.785} & \textbf{0.788} & 0.784 & 0.785 & 0.679 & 0.687 & 0.514 & 0.325 & 0.112 & 0.017 \\
AllGestureWiimoteY & 0.774 & 0.774 & 0.148 & 0.085 & 0.777 & 0.779 & \textbf{0.789} & \textbf{0.789} & 0.676 & 0.694 & 0.546 & 0.335 & 0.108 & 0.022 \\
AllGestureWiimoteZ & \textbf{0.774} & \textbf{0.782} & 0.132 & 0.058 & 0.759 & 0.761 & 0.754 & 0.756 & 0.649 & 0.662 & 0.548 & 0.344 & 0.125 & 0.032 \\
BME & 0.991 & 0.991 & \textbf{0.998} & \textbf{0.998} & 0.980 & 0.981 & 0.971 & 0.971 & 0.847 & 0.844 & 0.880 & 0.779 & 0.467 & 0.449 \\
Chinatown & 0.937 & 0.906 & \textbf{0.975} & \textbf{0.960} & 0.972 & 0.955 & 0.943 & 0.909 & 0.971 & 0.952 & 0.855 & 0.743 & 0.717 & 0.575 \\
Crop & \textbf{0.775} & \textbf{0.777} & 0.754 & 0.754 & 0.753 & 0.752 & 0.731 & 0.732 & 0.659 & 0.668 & 0.634 & 0.448 & 0.370 & 0.350 \\
EOGHorizontalSignal & \textbf{0.677} & \textbf{0.710} & 0.512 & 0.559 & 0.537 & 0.589 & 0.532 & 0.537 & 0.434 & 0.486 & 0.428 & 0.348 & 0.188 & 0.105 \\
EOGVerticalSignal & 0.461 & 0.496 & 0.397 & 0.437 & \textbf{0.484} & \textbf{0.508} & 0.453 & 0.457 & 0.436 & 0.448 & 0.379 & 0.286 & 0.173 & 0.080 \\
EthanolLevel & \textbf{0.808} & \textbf{0.815} & 0.582 & 0.583 & 0.474 & 0.470 & 0.315 & 0.316 & 0.280 & 0.273 & 0.322 & 0.265 & 0.253 & 0.125 \\
FreezerRegularTrain & \textbf{0.994} & \textbf{0.994} & 0.987 & 0.988 & 0.983 & 0.984 & 0.979 & 0.961 & 0.839 & 0.839 & 0.894 & 0.715 & 0.705 & 0.719 \\
FreezerSmallTrain & 0.781 & 0.783 & \textbf{0.901} & \textbf{0.904} & 0.870 & 0.870 & 0.883 & 0.864 & 0.760 & 0.761 & 0.692 & 0.543 & 0.593 & 0.624 \\
Fungi & 0.855 & 0.897 & 0.887 & 0.936 & 0.938 & 0.952 & \textbf{0.990} & \textbf{0.995} & 0.177 & 0.032 & 0.444 & 0.375 & 0.084 & 0.009 \\
GestureMidAirD1 & \textbf{0.726} & \textbf{0.752} & 0.091 & 0.029 & 0.623 & 0.678 & 0.562 & 0.565 & 0.438 & 0.466 & 0.624 & 0.490 & 0.052 & 0.004 \\
GestureMidAirD2 & \textbf{0.605} & \textbf{0.645} & 0.065 & 0.032 & 0.534 & 0.556 & 0.482 & 0.485 & 0.362 & 0.368 & 0.512 & 0.373 & 0.054 & 0.004 \\
GestureMidAirD3 & 0.266 & 0.270 & 0.075 & 0.033 & 0.312 & \textbf{0.315} & 0.277 & 0.271 & 0.169 & 0.099 & \textbf{0.330} & 0.231 & 0.051 & 0.020 \\
GesturePebbleZ1 & 0.865 & 0.879 & 0.228 & 0.149 & 0.852 & 0.878 & \textbf{0.884} & \textbf{0.880} & 0.657 & 0.692 & 0.752 & 0.728 & 0.161 & 0.027 \\
GesturePebbleZ2 & 0.851 & \textbf{0.860} & 0.209 & 0.098 & \textbf{0.857} & \textbf{0.860} & 0.843 & 0.837 & 0.582 & 0.653 & 0.727 & 0.700 & 0.155 & 0.037 \\
GunPointAgeSpan & \textbf{0.996} & \textbf{0.996} & 0.967 & 0.969 & 0.980 & 0.980 & 0.983 & 0.974 & 0.994 & 0.994 & 0.804 & 0.755 & 0.558 & 0.536 \\
GunPointMaleVersusFemale & 0.999 & 0.999 & 0.999 & 0.999 & \textbf{1.000} & \textbf{1.000} & \textbf{1.000} & 0.997 & 0.968 & 0.968 & 0.974 & 0.919 & 0.728 & 0.782 \\
GunPointOldVersusYoung & \textbf{1.000} & \textbf{1.000} & \textbf{1.000} & \textbf{1.000} & \textbf{1.000} & \textbf{1.000} & \textbf{1.000} & \textbf{1.000} & \textbf{1.000} & \textbf{1.000} & 0.999 & 0.998 & 0.840 & 0.878 \\
HouseTwenty & \textbf{0.943} & \textbf{0.941} & 0.846 & 0.851 & 0.902 & 0.910 & 0.933 & 0.843 & 0.882 & 0.883 & 0.804 & 0.620 & 0.578 & 0.654 \\
InsectEPGRegularTrain & \textbf{1.000} & \textbf{1.000} & \textbf{1.000} & \textbf{1.000} & \textbf{1.000} & \textbf{1.000} & \textbf{1.000} & \textbf{1.000} & \textbf{1.000} & \textbf{1.000} & \textbf{1.000} & \textbf{1.000} & 0.831 & 0.579 \\
InsectEPGSmallTrain & \textbf{1.000} & \textbf{1.000} & \textbf{1.000} & \textbf{1.000} & \textbf{1.000} & \textbf{1.000} & \textbf{1.000} & \textbf{1.000} & \textbf{1.000} & \textbf{1.000} & 0.916 & 0.911 & 0.714 & 0.548 \\
MelbournePedestrian & \textbf{0.958} & \textbf{0.958} & 0.928 & 0.933 & 0.955 & 0.955 & 0.938 & 0.938 & 0.831 & 0.849 & 0.807 & 0.672 & 0.253 & 0.200 \\
MixedShapesRegularTrain & \textbf{0.951} & \textbf{0.947} & 0.916 & 0.915 & 0.919 & 0.919 & 0.922 & 0.923 & 0.823 & 0.819 & 0.851 & 0.670 & 0.254 & 0.278 \\
MixedShapesSmallTrain & \textbf{0.878} & 0.872 & 0.831 & 0.839 & 0.852 & 0.848 & 0.872 & \textbf{0.877} & 0.739 & 0.748 & 0.789 & 0.602 & 0.240 & 0.239 \\
PickupGestureWiimoteZ & 0.800 & 0.836 & 0.142 & 0.072 & \textbf{0.840} & \textbf{0.868} & 0.804 & 0.808 & 0.680 & 0.687 & 0.739 & 0.610 & 0.140 & 0.028 \\
PigAirwayPressure & 0.533 & 0.577 & 0.205 & 0.258 & \textbf{0.626} & \textbf{0.656} & 0.465 & 0.464 & 0.053 & 0.041 & 0.088 & 0.073 & 0.031 & 0.002 \\
PigArtPressure & \textbf{0.978} & \textbf{0.983} & 0.444 & 0.463 & 0.964 & 0.972 & 0.924 & 0.922 & 0.322 & 0.278 & 0.164 & 0.143 & 0.040 & 0.008 \\
PigCVP & 0.797 & \textbf{0.834} & 0.192 & 0.196 & \textbf{0.811} & 0.821 & 0.740 & 0.738 & 0.202 & 0.198 & 0.119 & 0.102 & 0.040 & 0.005 \\
PLAID & 0.495 & 0.584 & 0.087 & 0.084 & 0.544 & 0.590 & 0.636 & 0.556 & \textbf{0.892} & \textbf{0.909} & 0.324 & 0.276 & 0.205 & 0.048 \\
PowerCons & 0.951 & 0.952 & 0.984 & 0.984 & 0.964 & 0.965 & 0.931 & 0.871 & 0.856 & 0.857 & \textbf{0.997} & \textbf{0.996} & 0.673 & 0.691 \\
Rock & 0.428 & 0.478 & 0.590 & 0.513 & \textbf{0.712} & \textbf{0.691} & 0.576 & 0.639 & 0.440 & 0.379 & 0.514 & 0.393 & 0.288 & 0.151 \\
SemgHandGenderCh2 & 0.665 & 0.749 & 0.870 & 0.856 & \textbf{0.956} & \textbf{0.949} & 0.908 & 0.861 & 0.912 & 0.907 & 0.919 & 0.805 & 0.588 & 0.640 \\
SemgHandMovementCh2 & 0.720 & 0.723 & 0.667 & 0.674 & \textbf{0.870} & \textbf{0.873} & 0.783 & 0.784 & 0.764 & 0.780 & 0.670 & 0.443 & 0.290 & 0.278 \\
SemgHandSubjectCh2 & 0.876 & 0.883 & 0.852 & 0.854 & \textbf{0.948} & \textbf{0.949} & 0.859 & 0.861 & 0.849 & 0.855 & 0.883 & 0.715 & 0.439 & 0.413 \\
ShakeGestureWiimoteZ & 0.880 & 0.901 & 0.182 & 0.077 & 0.924 & \textbf{0.939} & \textbf{0.928} & 0.929 & 0.820 & 0.779 & 0.807 & 0.779 & 0.104 & 0.015 \\
SmoothSubspace & 0.975 & 0.975 & 0.824 & 0.805 & \textbf{0.976} & \textbf{0.976} & 0.943 & 0.943 & 0.860 & 0.862 & 0.926 & 0.768 & 0.788 & 0.788 \\
UMD & 0.990 & 0.990 & 0.990 & 0.991 & \textbf{0.994} & \textbf{0.994} & 0.992 & 0.992 & 0.854 & 0.854 & 0.932 & 0.878 & 0.407 & 0.236 \\
DodgerLoopDay & 0.537 & 0.581 & \textbf{0.556} & \textbf{0.587} & 0.495 & 0.540 & -- & -- & 0.362 & 0.453 & 0.555 & 0.376 & 0.200 & 0.133 \\
DodgerLoopGame & 0.803 & 0.803 & 0.777 & 0.784 & 0.813 & 0.818 & -- & -- & \textbf{0.862} & \textbf{0.896} & 0.750 & 0.750 & 0.571 & 0.424 \\
DodgerLoopWeekend & 0.896 & 0.858 & \textbf{0.972} & \textbf{0.960} & 0.952 & 0.936 & -- & -- & 0.949 & 0.931 & 0.910 & 0.765 & 0.717 & 0.441 \\
AVG & 0.821 & \textbf{0.817} & 0.733 & 0.723 & \textbf{0.822} & 0.816 & 0.782 & 0.750 & 0.699 & 0.692 & 0.685 & 0.603 & 0.406 & 0.305  \\
\bottomrule
 \caption{Accuracy scores and Precision scores of our method compared with those of other methods on 128 UCR
datasets. The representation dimensions of TS2Vec, T-Loss, TNC, TS-TCC and TST are all set to 320 for fair comparison.}\label{tab4}  \\

\end{longtable}

\setlength{\tabcolsep}{4pt}
\begin{longtable}{lcccccccccccccc}
\hline
Dataset & \multicolumn{2}{c}{RankSCL} & \multicolumn{2}{c}{InfoTS} & \multicolumn{2}{c}{TS2Vec} & \multicolumn{2}{c}{T-Loss} & \multicolumn{2}{c}{DTW} & \multicolumn{2}{c}{TS-TCC} & \multicolumn{2}{c}{TNC} \\
\cmidrule(r){2-3}  \cmidrule(r){4-5} \cmidrule(r){6-7}  \cmidrule(r){8-9}  \cmidrule(r){10-11}  \cmidrule(r){12-13}  \cmidrule(r){14-15} 
& ACC & Prec. & ACC & Prec. & ACC & Prec. & ACC & Prec. & ACC & Prec. & ACC & Prec. & ACC & Prec. \\ \hline
\endfirsthead

\multicolumn{15}{c} 
{{\bfseries Table \thetable\ continued from previous page}} \\
\hline
Dataset & \multicolumn{2}{c}{RankSCL} & \multicolumn{2}{c}{InfoTS} & \multicolumn{2}{c}{TS2Vec} & \multicolumn{2}{c}{T-Loss} & \multicolumn{2}{c}{DTW} & \multicolumn{2}{c}{TS-TCC} & \multicolumn{2}{c}{TNC} \\ \cmidrule(r){2-3}  \cmidrule(r){4-5} \cmidrule(r){6-7}  \cmidrule(r){8-9}  \cmidrule(r){10-11}  \cmidrule(r){12-13}  \cmidrule(r){14-15} 
& ACC & Prec. & ACC & Prec. & ACC & Prec. & ACC & Prec. & ACC & Prec. & ACC & Prec. & ACC & Prec. \\ \hline
\endhead

\hline 
\multicolumn{15}{r}{{Continued on next page}} \\ \hline
\endfoot

\endlastfoot
       ArticularyWordRecognition & 0.977 & 0.978 & 0.982 & 0.983 & 0.981 & 0.984 & 0.961 & 0.961 & \textbf{0.983} & \textbf{0.985} & 0.927 & 0.869 & 0.579 & 0.601 \\
AtrialFibrillation & 0.293 & 0.309 & 0.141 & 0.121 & 0.240 & 0.395 & 0.133 & 0.133 & \textbf{0.467} & \textbf{0.435} & 0.304 & 0.228 & 0.281 & 0.165 \\
BasicMotions & \textbf{1.000} & \textbf{1.000} & 0.985 & 0.986 & 0.975 & 0.977 & 0.955 & 0.955 & 0.775 & 0.882 & 0.945 & 0.903 & 0.500 & 0.523 \\
CharacterTrajectories & \textbf{0.996} & \textbf{0.995} & 0.682 & 0.808 & 0.993 & 0.993 & 0.060 & 0.050 & 0.982 & 0.981 & 0.979 & 0.961 & 0.197 & 0.158 \\
Cricket & 0.986 & 0.987 & 0.971 & 0.974 & 0.972 & 0.975 & 0.989 & 0.989 & \textbf{1.000} & \textbf{1.000} & 0.778 & 0.721 & 0.553 & 0.485 \\
DuckDuckGeese & \textbf{0.604} & \textbf{0.642} & 0.493 & 0.543 & 0.496 & 0.495 & 0.580 & 0.579 & 0.340 & 0.296 & 0.331 & 0.295 & 0.356 & 0.369 \\
EigenWorms & 0.727 & 0.786 & 0.660 & 0.661 & 0.841 & \textbf{0.791} & \textbf{0.846} & 0.790 & 0.588 & 0.474 & 0.311 & 0.320 & 0.369 & 0.108 \\
Epilepsy & 0.975 & 0.977 & 0.959 & 0.958 & 0.963 & 0.961 & \textbf{0.978} & \textbf{0.979} & 0.935 & 0.938 & 0.859 & 0.815 & 0.317 & 0.284 \\
ERing & 0.913 & 0.915 & 0.932 & \textbf{0.937} & 0.850 & 0.870 & 0.838 & 0.837 & \textbf{0.933} & 0.935 & 0.884 & 0.850 & 0.501 & 0.519 \\
EthanolConcentration & 0.304 & \textbf{0.309} & 0.258 & 0.258 & 0.282 & 0.285 & 0.234 & 0.231 & 0.289 & 0.302 & 0.291 & 0.283 & \textbf{0.308} & 0.199 \\
FaceDetection & \textbf{0.637} & \textbf{0.638} & 0.520 & 0.520 & 0.512 & 0.512 & 0.518 & 0.505 & 0.534 & 0.534 & 0.540 & 0.506 & 0.518 & 0.523 \\
FingerMovements & 0.528 & 0.529 & 0.483 & 0.484 & 0.480 & 0.480 & 0.536 & 0.533 & \textbf{0.620} & \textbf{0.620} & 0.476 & 0.458 & 0.476 & 0.473 \\
HandMovementDirection & 0.322 & 0.333 & 0.359 & 0.362 & 0.308 & 0.340 & 0.305 & 0.334 & 0.189 & 0.268 & \textbf{0.434} & \textbf{0.373} & 0.286 & 0.247 \\
Handwriting & \textbf{0.566} & \textbf{0.560} & 0.394 & 0.418 & 0.520 & 0.536 & 0.425 & 0.438 & 0.500 & 0.548 & 0.297 & 0.204 & 0.061 & 0.029 \\
Heartbeat & 0.740 & 0.678 & \textbf{0.745} & 0.710 & 0.711 & 0.675 & 0.713 & 0.647 & 0.722 & 0.630 & 0.724 & \textbf{0.717} & 0.694 & 0.589 \\
JapaneseVowels & 0.980 & 0.977 & \textbf{0.988} & \textbf{0.988} & 0.985 & 0.987 & 0.080 & 0.110 & 0.968 & 0.968 & 0.858 & 0.785 & 0.189 & 0.179 \\
Libras & \textbf{0.914} & \textbf{0.923} & 0.849 & 0.856 & 0.851 & 0.860 & 0.868 & 0.868 & 0.867 & 0.868 & 0.701 & 0.641 & 0.250 & 0.202 \\
LSST & 0.164 & 0.201 & \textbf{0.566} & \textbf{0.454} & 0.543 & 0.416 & 0.528 & 0.325 & 0.559 & 0.421 & 0.366 & 0.379 & 0.360 & 0.216 \\
MotorImagery & \textbf{0.550} & \textbf{0.557} & 0.472 & 0.472 & 0.498 & 0.249 & 0.526 & 0.500 & 0.490 & 0.490 & 0.526 & 0.418 & 0.534 & 0.536 \\
NATOPS & \textbf{0.927} & \textbf{0.928} & 0.922 & 0.924 & 0.916 & 0.916 & 0.912 & 0.912 & 0.878 & 0.877 & 0.729 & 0.631 & 0.301 & 0.341 \\
PEMS-SF & \textbf{0.881} & \textbf{0.886} & 0.705 & 0.714 & 0.684 & 0.692 & 0.635 & 0.638 & 0.555 & 0.564 & 0.676 & 0.570 & 0.301 & 0.302 \\
PenDigits & 0.986 & 0.987 & 0.987 & 0.987 & \textbf{0.989} & \textbf{0.989} & 0.986 & 0.986 & 0.979 & 0.979 & 0.910 & 0.871 & -- & -- \\
PhonemeSpectra & \textbf{0.271} & \textbf{0.271} & 0.209 & 0.209 & 0.234 & 0.233 & 0.231 & 0.232 & 0.164 & 0.177 & 0.155 & 0.140 & 0.065 & 0.050 \\
RacketSports & \textbf{0.881} & \textbf{0.891} & 0.832 & 0.845 & 0.850 & 0.869 & 0.810 & 0.824 & 0.822 & 0.838 & 0.794 & 0.753 & 0.408 & 0.423 \\
SelfRegulationSCP1 & 0.862 & 0.867 & \textbf{0.867} & \textbf{0.870} & 0.786 & 0.815 & 0.832 & 0.866 & 0.833 & 0.850 & 0.862 & 0.850 & 0.627 & 0.666 \\
SelfRegulationSCP2 & 0.549 & 0.549 & 0.536 & 0.539 & \textbf{0.560} & \textbf{0.560} & 0.541 & 0.525 & 0.517 & 0.517 & 0.504 & 0.461 & 0.518 & 0.517 \\
SpokenArabicDigits & \textbf{0.989} & \textbf{0.989} & 0.915 & 0.921 & 0.988 & 0.988 & 0.100 & 0.100 & 0.980 & 0.981 & 0.959 & 0.941 & 0.123 & 0.060 \\
StandWalkJump & \textbf{0.533} & \textbf{0.529} & 0.403 & 0.400 & 0.480 & 0.418 & 0.320 & 0.320 & 0.267 & 0.103 & 0.379 & 0.357 & 0.320 & 0.227 \\
UWaveGestureLibrary & 0.873 & 0.874 & 0.780 & 0.780 & \textbf{0.898} & 0.901 & 0.893 & 0.892 & 0.897 & \textbf{0.903} & 0.755 & 0.650 & 0.351 & 0.328 \\
InsectWingbeat & \textbf{0.518} & \textbf{0.512} & 0.464 & 0.468 & 0.464 & 0.460 & 0.100 & 0.100 & -- & -- & 0.253 & 0.225 & -- & -- \\
AVG & \textbf{0.715} & \textbf{0.719} & 0.669 & 0.672 & 0.695 & 0.687 & 0.581 & 0.572 & 0.654 & 0.645 & 0.617 & 0.573 & 0.345 & 0.311 \\

\bottomrule
\caption{Accuracy scores and Precision scores of our method compared with those of other methods on 30 UEA
datasets. }\label{tab5}  \\
\end{longtable}

}

\end{document}